\documentclass[twocolumn]{camel-ai}

\usepackage[toc,page,header]{appendix}

\usepackage[utf8]{inputenc} 
\usepackage[T1]{fontenc}    
\usepackage{hyperref}       
\usepackage{url}            
\usepackage{booktabs}       
\usepackage{amsfonts}       
\usepackage{nicefrac}       
\usepackage{microtype}      
\usepackage{xspace}
\usepackage{markdown}
\usepackage{bm}
\usepackage{bbm}
\usepackage{tabularx}
\usepackage{amsmath}
\usepackage{mathtools}
\usepackage{amsthm}
\usepackage{pifont} 
\usepackage{multirow}
\usepackage{makecell}
\usepackage{paralist}
\usepackage{color}
\usepackage{colortbl}
\usepackage{adjustbox}
\usepackage[edges]{forest}
\usepackage{amssymb}
\usepackage{pifont}
\usepackage{wrapfig}
\usepackage[noend]{algpseudocode}
\usepackage{algorithm}
\usepackage{algorithmicx}
\usepackage{algpseudocode}
\usepackage{pifont}
\usepackage{pgfplots}
\pgfplotsset{compat=1.17}
\usepackage{pgfplotstable}
\usepackage{svg}
\usepgfplotslibrary{groupplots}
\usepackage{microtype}
\usepackage{pgf-pie}
\usepackage{graphicx}
\usepackage{booktabs} 
\usepackage{multirow} 
\usepackage[utf8]{inputenc}
\usepackage[most]{tcolorbox}
\usepackage{markdown}
\usepackage{pgfplots}     
\pgfplotsset{compat=1.17}
\usepackage{siunitx}
\usepackage{enumitem}
\usepackage{amssymb}

\usepackage{caption}
\usepackage{graphicx}
\usepackage[most]{tcolorbox}
\newtcolorbox{mycustombox}[1]{
  enhanced,
  colback=black!5!white,
  colframe=black!75!white,
  boxrule=0.4pt,
  coltitle=white,
  title=#1,
  titlerule=0.4pt,
  fontupper=\small,
  fonttitle=\small,
  before upper={\par\smallskipamount},
  breakable,
  left=3mm,
  right=3mm,
  top=2mm,
  bottom=2mm,
  boxsep=1mm,
}

\definecolor{mygray}{gray}{0.9}
\definecolor{newgreen}{RGB}{78, 173, 102}
\definecolor{speciallink}{HTML}{5a2afd}

\newtheorem{remark}{Remark}
\definecolor{mycolor_green}{HTML}{E6F8E0}
\definecolor{mmada_color}{HTML}{EFF7FF}
\definecolor{mycolor_blue}{HTML}{E7EFFA}
\definecolor{mycolor_green}{HTML}{E6F8E0}
\definecolor{mycolor_gray}{HTML}{ECECEC}
\definecolor{pearDark}{HTML}{2980B9}
\definecolor{demphcolor}{RGB}{90,42,253}
\definecolor{mydarkgreen}{RGB}{0,100,0}

\newcommand{\redcross}{\tikz[baseline=-0.5ex]{\draw[red, thick] (-0.2ex,-0.2ex) -- (0.8ex,0.8ex); \draw[red, thick] (-0.2ex,0.8ex) -- (0.8ex,-0.2ex);}}
\newcommand{\greentick}{%
  \tikz[baseline=-0.5ex, line width=0.8pt]{
    \draw[green!60!black] (0ex,0.5ex) -- (0.4ex,0ex) -- (1ex,1.2ex);
  }
}
\usepackage{textcomp}  
\usepackage{scalerel}  

\usepackage{times}

\usepackage[utf8]{inputenc}
\usepackage{microtype}
\usepackage{graphicx}
\usepackage{booktabs}
\usepackage{hyperref}
\usepackage{stackengine}
\usepackage{wrapfig}
\usepackage{etoolbox}
\usepackage{fancyvrb}
\usepackage{multirow}
\usepackage{amssymb}
\usepackage{adjustbox}
\usepackage{algorithm}
\usepackage{algorithmicx}
\usepackage{algpseudocode}
\usepackage{amsfonts}       
\usepackage{booktabs}       
\usepackage{float}
\usepackage{fnpos}
\usepackage{graphicx}
\usepackage{microtype}      
\usepackage{nicefrac}       
\usepackage[textsize=footnotesize]{todonotes}
\usepackage{xspace}
\usepackage{mathtools}
\usepackage{enumitem}
\usepackage{thmtools}
\usepackage{wrapfig}
\usepackage{bbm}
\usepackage{wrapfig}
\usepackage{transparent}
\usepackage{enumitem}
\usepackage{changepage} 
\usepackage{pythonhighlight}
\usepackage{xurl}
\usepackage{subcaption}
\usepackage{svg}
\usepackage{booktabs}

\usepackage[most]{tcolorbox}
\usetikzlibrary{shadows}
\definecolor{mine}{RGB}{205, 232, 248}%
\definecolor{minedark}{RGB}{160, 190, 210}%
\definecolor{revision}{RGB}{210, 22, 123}

\usepackage[]{nomencl}   
    \makenomenclature

\makeatletter
\newcommand\myscriptsize{\@setfontsize\myscriptsize{8pt}{9pt}}
\makeatother

\providetoggle{nomsort}
\settoggle{nomsort}{true} 

\makeatletter
\iftoggle{nomsort}{%
    \let\old@@@nomenclature=\@@@nomenclature        
        \newcounter{@nomcount} \setcounter{@nomcount}{0}%
        \renewcommand\the@nomcount{\two@digits{\value{@nomcount}}}
        \def\@@@nomenclature[#1]#2#3{
          \addtocounter{@nomcount}{1}%
        \def\@tempa{#2}\def\@tempb{#3}%
          \protected@write\@nomenclaturefile{}%
          {\string\nomenclatureentry{\the@nomcount\nom@verb\@tempa @[{\nom@verb\@tempa}]%
          \begingroup\nom@verb\@tempb\protect\nomeqref{\theequation}%
          |nompageref}{\thepage}}%
          \endgroup
          \@esphack}%
      }{}
\makeatother
\usepackage{silence}
\usepackage{tikz}

\tikzstyle{every picture}+=[remember picture]

\usepackage[most]{tcolorbox}
\usetikzlibrary{shadows}

\definecolor{functionclass}{RGB}{113, 153, 194}
\definecolor{lossfunction}{RGB}{119, 221, 119}
\definecolor{regterm}{RGB}{255, 179, 71}

\newcounter{exa}
\definecolor{gblue}{RGB}{66,133,244}
\definecolor{gred}{RGB}{219,68,55}
\definecolor{gyellow}{RGB}{244,180,0}
\definecolor{ggreen}{RGB}{15,157,88}
\definecolor{lpcolor}{RGB}{42,74,138}
\definecolor{morelcolor}{RGB}{185,18,32}
\definecolor{bgcolor}{RGB}{230,245,208}
\definecolor{framecolor}{RGB}{244,109,67}
\definecolor{mulberry}{rgb}{0.77, 0.29, 0.55}
\hypersetup{colorlinks=true,
            citecolor=gblue,
            urlcolor=ggreen,
            linkcolor=purple}
\definecolor{Chocolate3}{RGB}{205, 105, 29}
\definecolor{LightSteelBlue3}{RGB}{162, 181, 205}
\definecolor{DodgerBlue4}{RGB}{16, 78, 139}

\tcbset{
myexample/.style={
  enhanced,
  colback=yellow!10!white,
  colframe=red!50!black,
  fonttitle=\scshape,
  titlerule=0pt,
  title={\refstepcounter{exa}example~\theexa.},
  title style={fill=yellow!10!white},
  coltitle=red!50!black,
  drop shadow,
  highlight math style={reset,colback=LightBlue!50!white,colframe=Navy}
  }
  }
\newtcolorbox{texample}{myexample}

\newtheorem{exampp}{Example}
\usepackage{framed}
\colorlet{shadecolor}{gray!20}

\colorlet{LightLavender}{green!5}
\tcbset{on line, 
        boxsep=4pt, left=0pt,right=0pt,top=0pt,bottom=0pt,
        colframe=white,colback=LightLavender,  
        highlight math style={enhanced}
        }

\allowdisplaybreaks

\usepackage{mathtools}
\usepackage{natbib}
\usepackage{latexsym}

\usepackage{url}
\usepackage{amssymb}
\usepackage[utf8]{inputenc}
\usepackage{microtype}
\usepackage{booktabs}
\usepackage{pifont} 
\usepackage{multirow}
\usepackage{makecell}
\usepackage{paralist}
\usepackage{xspace}
\usepackage{color}
\usepackage{colortbl}
\usepackage{adjustbox}
\usepackage{hyperref} 
\usepackage[edges]{forest}
\usepackage{tikz} 
\usepackage{caption}
\usepackage{amsfonts}

\hypersetup{
    colorlinks,
    linkcolor={blue!80!black},
    citecolor={blue!80!black},
}
\tikzset{
    root/.style =             {align=center, text width=1cm, rounded corners=3pt, line width=0.3mm, fill=gray!10, draw=gray!80, font=\small},
    demographic/.style =         {align=center, text width=1.8cm, rounded corners=3pt, line width=0.3mm, fill=blue!10, draw=blue!80, font=\footnotesize},
    demographic_work/.style =    {align=center, text width=10cm, rounded corners=3pt, line width=0.3mm, fill=blue!10, draw=blue!0, font=\footnotesize},
    character/.style =         {align=center, text width=1.8cm, rounded corners=3pt, line width=0.3mm, fill=red!10, draw=red!80, font=\footnotesize},
    character_work/.style =    {align=center, text width=10cm, rounded corners=3pt, line width=0.3mm, fill=red!10, draw=red!0, font=\footnotesize},
    personalization/.style =           {align=center, text width=1.8cm, rounded corners=3pt, line width=0.3mm, fill=cyan!10, draw=cyan!80, font=\footnotesize},
    personalization_work/.style =      {align=center, text width=10cm, rounded corners=3pt, line width=0.3mm, fill=cyan!10, draw=cyan!0, font=\footnotesize},
    risk/.style =         {align=center, text width=1.8cm, rounded corners=3pt, line width=0.3mm, fill=orange!10, draw=orange!80, font=\footnotesize},
    risk_work/.style =    {align=center, text width=10cm, rounded corners=3pt, line width=0.3mm, fill=orange!10, draw=orange!0, font=\footnotesize},
}

\usepackage{CJK}
\definecolor{myblue}{RGB}{34, 66, 122} 
\renewcommand{\eqref}[1]{Eq.~\ref{#1}}

\usepackage[toc,page,header]{appendix}
\usepackage{minitoc}

\usepackage{amsmath,amsfonts,bm,booktabs}
\usepackage{multicol,multirow}

\def\eqref#1{equation~\ref{#1}}

\def\1{\bm{1}}

\DeclareMathAlphabet{\mathsfit}{\encodingdefault}{\sfdefault}{m}{sl}
\SetMathAlphabet{\mathsfit}{bold}{\encodingdefault}{\sfdefault}{bx}{n}

\title{Distilling Tool Knowledge into Language Models via Back-Translated Traces}

\author[1,2,3*$\S$]{Xingyue Huang}
\author[4*]{Xianglong Hu}
\author[2,3,5*]{Zifeng Ding}
\author[4,1,2$\dagger$]{Yuan He}
\author[2,3$\S$]{Rishabh}
\author[2,6]{Waleed Alzarooni}
\author[7]{Ziyu Ye}
\author[2,3]{Wendong Fan}
\author[8]{Bailan He}
\author[9]{Haige Bo}
\author[9]{Changran Hu}
\author[2,3]{Guohao Li}

\affiliation[1]{University of Oxford}
\affiliation[2]{CAMEL-AI}
\affiliation[3]{Eigent.AI}
\affiliation[4]{Amazon}
\affiliation[5]{University of Cambridge}
\affiliation[6]{University of Los Angeles}
\affiliation[7]{University of Chicago}
\affiliation[8]{Ludwig Maximilian University of Munich}
\affiliation[9]{SambaNova Systems}

\contribution[*]{Equal Contribution}

\contribution[\S]{Work done during internship at Eigent.AI}
\contribution[\dagger]{Work done prior to joining Amazon}

\abstract{
Large language models (LLMs) often struggle with mathematical problems that require exact computation or multi-step algebraic reasoning. Tool-integrated reasoning (TIR) offers a promising solution by leveraging external tools such as code interpreters to ensure correctness, but it introduces inference-time dependencies that hinder scalability and deployment. In this work, we propose a new paradigm for distilling tool knowledge into LLMs purely through natural language. We first construct a \textsc{Solver Agent} that solves math problems by interleaving planning, symbolic tool calls, and reflective reasoning. Then, using a back-translation pipeline powered by multiple LLM-based agents, we convert interleaved TIR traces into natural language reasoning traces. A \textsc{Translator Agent} generates explanations for individual tool calls, while a \textsc{Rephrase Agent} merges them into a fluent and globally coherent narrative. Empirically, we show that fine-tuning a small open-source model on these synthesized traces enables it to internalize both tool knowledge and structured reasoning patterns, yielding gains on competition-level math benchmarks without requiring tool access at inference.
}

\date{June 24, 2025}
\correspondence{Guohao Li at \href{mailto:guohao.li@eigent.ai}{\textcolor{speciallink}{guohao.li@eigent.ai}}}

\begin{document}
\maketitle

\section{Introduction}
Mathematical reasoning remains a core challenge for large language models (LLMs), particularly due to their limitations in executing precise computations and handling symbolic structures. While recent advances in chain-of-thought (CoT) prompting~\citep{cot} have improved performance on multi-step reasoning tasks, pure natural language (NL) solutions often falter on problems that require exact arithmetic, algebraic manipulation, or symbolic calculus.

To overcome these limitations, a growing body of work~\citep{DBLP:journals/tmlr/ChenM0C23,DBLP:conf/emnlp/YinYJ0B24,DBLP:conf/iclr/GouSGSYHDC24,goldie2025syntheticdatageneration,li2025torl} has explored tool-integrated reasoning (TIR), where models augment their reasoning with calls to external symbolic tools such as Python Interpreter coupled with libraries targeted on accurate mathematical calculations, e.g., \texttt{SymPy}. These methods offer clear benefits: tool calls ensure arithmetic precision, improve intermediate consistency, and enable modular problem decomposition. However, the reliance on tool access at inference time creates deployment frictions and complicates the use of such models in restricted environments. Moreover, it remains unclear whether models can follow the structured reasoning patterns enabled by tool-integrated approaches while relying solely on NL without invoking any external tools. To this end, we aim to ask the following question:

\textit{Can we transfer the competence of tool-integrated agents into language models?}

\paragraph{Approach.}  
In this work, we propose a new paradigm for transferring the benefits of TIR into standard language models without requiring tool access at inference. Our approach is centered around a back-translation pipeline empowered by multiple LLM-based agents that converts high-quality TIR traces, generated via a \texttt{SymPy}-based tool-calling agent, into pure NL-based reasoning traces. This enables supervised fine-tuning (SFT) on symbolic reasoning patterns, allowing models to internalize tool-driven problem-solving strategies while remaining tool-free at test time.

Specifically, we first design a \textsc{Solver Agent} that performs multi-step mathematical reasoning using a curated suite of symbolic tools. The agent generates structured traces by planning, invoking tools through the OpenAI function-calling interface, reflecting on outputs, and updating its strategy iteratively. These traces interleave tool calls with NL and are not immediately usable as language-only training examples. To overcome this, we develop a \textsc{Translator Agent} that translates each tool call into a fully self-contained tool call-specific NL reasoning trace. A corresponding \textsc{Judge Agent} verifies the correctness of each translation by comparing the answer provided by the translation to the original tool call output. Finally, we employ a \textsc{Rephrase Agent} to transform the entire trace, combining both high-level planning and computational steps, into a coherent, holistic NL-based reasoning trace. 

The resulting back-translated traces preserve the symbolic integrity of tool-augmented reasoning while making it accessible in a language-only format. This allows us to train open-source models through standard SFT, enabling them to mimic the behavior of tool-equipped agents on complex math tasks-without executing tools at inference.

\paragraph{Contribution.}
Our contribution can be summarized as follows. We conduct a systematic study of tool-integrated reasoning (TIR) in mathematical problem solving, leveraging a curated \texttt{SymPy} toolkit to enable symbolic computation within language models. To eliminate tool dependency at inference, we introduce a back-translation pipeline powered by multiple LLM-based agents that transforms interleaved tool-augmented traces into natural language (NL), enabling standard supervised fine-tuning. Empirically, we show that fine-tuning a small open-source model on these back-translated traces improves performance on challenging math benchmarks. We release our synthesized dataset of 11.6k NL reasoning traces derived from TIR examples to support future research.

\section{Related Work}\label{sec:lit}
\paragraph{Tool-Integrated Math Reasoning.}
Accurate calculation is essential for mathematical reasoning. Prior works show that relying solely on NL-based reasoning often yields suboptimal results, while incorporating tools, such as Python interpreters, can significantly improve performance. Early studies focus on single-step tool use, directly generating code to solve entire problems \citep{DBLP:journals/tmlr/ChenM0C23,DBLP:conf/icml/GaoMZ00YCN23,DBLP:conf/iclr/YueQZFH00C24}. More recent TIR frameworks interleave NL planning with program generation and execution across multiple steps, enabling more effective multi-step reasoning \citep{DBLP:conf/emnlp/YinYJ0B24,DBLP:conf/iclr/GouSGSYHDC24,DBLP:conf/iclr/WangRZLLSZSZ024,jia2024numinamath,goldie2025syntheticdatageneration,li2025torl}. Building on this interleaved paradigm, we propose an agent-based method that leverages a predefined set of tools, each implemented as a Python function with detailed usage descriptions, including input/output formats. Unlike prior TIR methods, our agent does not generate code but instead selects appropriate tools and inputs through language alone, lowering the reliance on code-writing skills while promoting coherent and flexible reasoning.

\paragraph{Synthetic Math Data Generation.}
Despite the availability of existing math datasets, the overall quantity of data with high-quality reasoning traces remains limited. Such data is crucial for improving LLMs' reasoning but is difficult to scale. Early synthetic data efforts, including popular works such as WizardMath \citep{DBLP:journals/corr/abs-2308-09583}, MetaMath \citep{DBLP:conf/iclr/YuJSYLZKLWL24}
and NuminaMath \citep{jia2024numinamath}, rely on frontier LLMs to directly generate traces, which limits quality when problems exceed the model's capability. Recent works start to explore more reliable methods for generating high-quality synthetic math data.
A recent work, rStar-Math \citep{DBLP:journals/corr/abs-2501-04519}, uses Monte Carlo Tree Search (MCTS) to generate and verify code-based reasoning traces. While effective, this method introduces significant computational overhead and produces traces in code without NL back-translation. Consequently, models trained on its data primarily learn to generate executable code rather than develop generalizable reasoning skills in NL. Another concurrent work SWiRL~\citep{goldie2025syntheticdatageneration} generates multi-step tool-augmented solution trajectories on math datasets. Its synthetic data omits the reasoning behind tool executions and is primarily designed to teach models effective tool usage, rather than to enhance their generic mathematical reasoning abilities. Unlike previous works, our approach collects TIR traces produced autonomously by the agent based on tool descriptions, and back-translates them into coherent NL solutions.

\paragraph{Post-Training on Synthetic Math Data.}
To improve LLMs' mathematical reasoning abilities, recent works have explored a variety of post-training strategies over synthetic math data. We discuss several notable works here. 
WizardMath \citep{DBLP:journals/corr/abs-2308-09583} trains on synthetic data generated via an evol-instruct pipeline, using reward models to evaluate instruction quality and step-by-step reasoning, followed by reinforcement learning (RL) to refine the model based on these signals. 
ToRA \citep{DBLP:conf/iclr/GouSGSYHDC24} is trained using imitation learning on curated tool-use trajectories, combined with output space shaping, to teach the model how to reason effectively with external tools. 
Numina-TIR \citep{jia2024numinamath} is trained first on NL-based CoTs, and then further fine-tuned on synthetic data that combines tool use (Python code) with reasoning. 
Concurrent works have shifted focus towards jointly training models to develop both mathematical reasoning and tool-use capabilities.
ToRL \citep{li2025torl} trains models using the GRPO~\citep{deepseek-math} algorithm with answer correctness reward, directly optimizing them on TIR traces.
rStar-Math \citep{DBLP:journals/corr/abs-2501-04519} alternately trains a policy model and a process preference model (PPM) using code-based MCTS trajectories selected via step-level Q-values.

\section{Generating TIR Traces}
\label{sec:sympy}

To solve math problems, we follow recent work~\citep{DBLP:conf/iclr/GouSGSYHDC24} by generating CoTs that interleave NL reasoning with structured tool calls. In contrast to methods that rely on free-form code generation and execution via a Python interpreter, we define a constrained set of tool functions within a curated \texttt{SymPy} toolkit. A dedicated \textsc{Solver Agent} orchestrates the reasoning process, deciding when and how to invoke these tools during problem solving.

\subsection{\texttt{SymPy} Toolkit Construction}

\begin{table}[t]
\centering
\scriptsize
\renewcommand{\arraystretch}{1.2}
\caption{Summary of Implemented Tools in our \texttt{Sympy} toolkit.}
\label{tab:symbolic_tools}
\begin{tabular}{ll}
\toprule
\textbf{Category} & \textbf{Tool Name (Function Name)} \\
\midrule
\multirow{3}{*}{Algebraic Simplification}
  & \texttt{simplify\_expression} \\
  & \texttt{expand\_expression} \\
  & \texttt{factor\_expression} \\
\midrule
\multirow{4}{*}{Equation Solving}
  & \texttt{solve\_equation} \\
  & \texttt{solve\_linear\_system} \\
  & \texttt{solve\_nonlinear\_system} \\
  & \texttt{find\_roots} \\
\midrule
\multirow{2}{*}{Inequalities}
  & \texttt{solve\_univariate\_inequality} \\
  & \texttt{reduce\_inequalities} \\
\midrule
\multirow{3}{*}{Polynomial Analysis}
  & \texttt{polynomial\_representation} \\
  & \texttt{polynomial\_degree} \\
  & \texttt{polynomial\_coefficients} \\
\midrule
\multirow{5}{*}{Calculus}
  & \texttt{differentiate} \\
  & \texttt{integrate} \\
  & \texttt{definite\_integral} \\
  & \texttt{series\_expansion} \\
  & \texttt{compute\_limit} \\
\midrule
\multirow{2}{*}{Critical Point Analysis}
  & \texttt{find\_critical\_points} \\
  & \texttt{check\_continuity} \\
\midrule
\multirow{7}{*}{Linear Algebra}
  & \texttt{compute\_determinant} \\
  & \texttt{compute\_inverse} \\
  & \texttt{compute\_eigenvalues} \\
  & \texttt{compute\_eigenvectors} \\
  & \texttt{compute\_nullspace} \\
  & \texttt{compute\_rank} \\
  & \texttt{compute\_inner\_product} \\
\bottomrule
\end{tabular}
\vspace{-2em}
\end{table}

To enable symbolic mathematical reasoning within language models, we construct a modular tool-calling interface built on top of \texttt{SymPy}~\citep{meurer2017sympy}. The toolkit includes a set of callable functions that span core mathematical domains such as algebra, calculus, equation solving, and linear algebra (see \Cref{tab:symbolic_tools}). Each tool is implemented as a Python wrapper with a structured docstring that defines its purpose, input, and output, which we convert into function signatures compatible with function-calling interface. This allows the \textsc{Solver Agent} to invoke tools without writing code, relying solely on the structured interface for tool selection and usage. See full definition of the toolkits in \Cref{app:sympy_toolkit}.

Compared to direct code generation, which requires the model to produce syntactically correct code, solving math problems with multiple tool calls is more reliable and interpretable. Tool calls abstract away low-level execution, offer built-in input format validation as well as error handling, and naturally help divide problem solving into explicit, modular reasoning steps. 
This simplifies both execution and supervision, enabling language models to focus on high-level planning and symbolic manipulation rather than low-level implementation details.

\subsection{Solving Problems with \textsc{Solver Agent}}

Given a symbolic toolkit, the \textsc{Solver Agent} is designed to produce interleaved reasoning traces that combine NL with structured tool calls, which will go through two phases: high-level planning, and step-wise execution loop with tool use. 
For instance, consider the following question as a running example: 

\begin{tcolorbox}[colframe=black!50, colback=yellow!5, sharp corners=south, boxrule=0.4pt, left=6pt, right=6pt, top=4pt, bottom=4pt]
\small
Let ( $C_1$ ) and ( $C_2$ ) be circles of radius 1 that are in the same plane and tangent to each other. Determine how many circles of radius 3 can be placed in this plane such that they are tangent to both ( $C_1$ ) and ( $C_2$ ).
\end{tcolorbox}

\paragraph{Planning.} The solving process begins with a high-level planning phase, where the model is prompted to generate a step-by-step strategy, explicitly stating which tools will be used and why. This encourages structured decomposition and discourages premature tool usage. 

\begin{tcolorbox}[colframe=black!60, colback=gray!5, sharp corners=south,title=Planning Prompt]
\scriptsize
\begin{verbatim}
Read the following question and provide a 
high-level, step-by-step plan for this problem, 
including the tools you will use.  
For each tool, explain why you are using 
it and what you expect it to return.  
Keep the plan high-level and concise.
\end{verbatim}
\end{tcolorbox}

Below, we show the step-by-step plan generated by the model in response to the prompt above, applied to our example problem involving tangency between circles. To solve this problem, the \textsc{Solver Agent} first interprets the geometric configuration and then generates a plan to determine the set of all valid positions for a circle of radius 3 that is tangent to both \( C_1 \) and \( C_2 \). This involves identifying the conditions for tangency, formulating equations for circle centers, and solving for valid solutions using symbolic tools.

\begin{figure}
\begin{tcolorbox}[colframe=brown!80, colback=yellow!5, sharp corners=south, boxrule=0.4pt, left=6pt, right=6pt, top=4pt, bottom=4pt, title= Example Output for Planning]
\scriptsize
\textbf{Step-by-Step Plan.}
\begin{enumerate}[noitemsep, topsep=0pt]
    \item \textbf{Understand the problem setup:} We have two circles \( C_1 \) and \( C_2 \) of radius 1, tangent to each other. We want to find how many circles of radius 3 can be tangent to both \( C_1 \) and \( C_2 \).

    \item \textbf{Set up coordinate system and variables:} Place \( C_1 \) and \( C_2 \) in the coordinate plane for convenience. For example, place \( C_1 \) at the origin and \( C_2 \) at \((2, 0)\), since they are tangent and each has radius 1.

    \item \textbf{Express conditions for tangency:}
    \begin{itemize}[noitemsep, topsep=0pt]
        \item The circle of radius 3 must be tangent to both \( C_1 \) and \( C_2 \).
        \item Use the distance between centers condition: the distance between the center of the new circle and the center of \( C_1 \) must be either \(3 + 1 = 4\) (external tangency) or \(|3 - 1| = 2\) (internal tangency).
        \item Similarly for \( C_2 \).
    \end{itemize}

    \item \textbf{Formulate equations:}...

    \item \textbf{Solve the system of equations:} ...

    \item \textbf{Count the number of solutions:} ...
\end{enumerate}

\textbf{Tools to use.}
\begin{itemize}[noitemsep, topsep=0pt]
    \item \textbf{Coordinate geometry:} ...
    \item \textbf{Distance formula:} ...
    \item \textbf{Equation solving (algebraic):} ...
    \item \textbf{Geometric reasoning:} ...
\end{itemize}

We expect a finite number of valid solutions, each corresponding to a circle of radius 3 tangent to both \( C_1 \) and \( C_2 \).

\end{tcolorbox}
\end{figure}

\paragraph{Step-wise Execution with Tool Use.}  
After generating a high-level plan, the \textsc{Solver Agent} proceeds through an iterative step-by-step execution loop following the generated plan. In each step, it determines whether a tool call is required, formulates the input, and attempts to invoke the tool. For each tool call, it allows up to five retries in case of failure, each constrained by a 30-second timeout. Once a valid output is obtained, the agent reflects on the result, assesses whether it aligns with the current goal, and updates its plan accordingly. It then determines the next action-either continuing with NL reasoning or invoking another tool. The prompt to drive each step of \textsc{Solver Agent}'s execution is shown below. 

\begin{tcolorbox}[colframe=black!60, colback=gray!5, sharp corners=south, title=Step Execution Prompt]
\scriptsize
\begin{verbatim}
Execute the next step of the plan.
You should reflect on the execution result, refine
your plan, and output the next action.  
Not all steps require a tool call - decide based on 
the current state and available tools (e.g., SymPy).
\end{verbatim}
\end{tcolorbox}
  
To guide this process, we ask the model to produce (1) the tool execution output, (2) a reflection on its significance, (3) a refined plan, and (4) the next step to take. This loop continues until a final answer is produced, wrapped within \verb|\boxed{}|, or a predefined limit of 15 steps is reached. We present in the following one step of the execution trace (abbreviated due to page limit) for our running example, and we include the full reasoning trace of our running example in \Cref{sec:example_tir_trace}.

\begin{figure}
\begin{tcolorbox}
  [colframe=brown!80, colback=yellow!5, sharp corners=south, boxrule=0.4pt, left=6pt, right=6pt, top=4pt, bottom=4pt, title=Example Output for Step Execution]
\scriptsize
Step 1: Understand the problem setup.

We have two circles \( C_1 \) and \( C_2 \), each with radius 1, tangent to each other. We want to 
find how many circles of radius 3 can be tangent to both \( C_1 \) and \( C_2 \).

Next, I will proceed to Step 2: ...

\scriptsize
\begin{verbatim}
<tool>
<tool_name>solve_nonlinear_system</tool_name>
<args>{
  "sympy_equations": [
    "(x-0)**2 + (y-0)**2 - 4**2",
    "(x-2)**2 + (y-0)**2 - 4**2"
  ],
  "variables": ["x", "y"]
}</args>
<result>
  Tuple(Integer(1), Pow(Integer(15),
  Rational(1, 2))),
  Tuple(Integer(1), Mul(Integer(-1), 
  Pow(Integer(15), Rational(1, 2))))
</result>
</tool>
\end{verbatim}

\scriptsize
We successfully computed the center coordinates of a circle of radius 3 that is externally tangent to both \( C_1 \) and \( C_2 \) ... 

We now proceed to check additional tangency configurations ...
\end{tcolorbox}
\end{figure}

Additionally, to ensure correctness and consistency, the \textsc{Solver Agent} is guided by a detailed system prompt that enforces strict \texttt{SymPy} grammar, such as using \texttt{Abs(x)} instead of \texttt{|x|}, \texttt{Eq()} instead of \texttt{=}, and proper handling of vectors, coordinates, and inner products. The prompt also encourages disciplined, context-aware tool usage and discourages unnecessary or invalid invocations. Full details are provided in \Cref{app:system_prompt}.

\subsection{Quantitative Analysis of TIR}  
We evaluate the effectiveness of our toolkit-based TIR framework on six advanced categories from the MATH dataset~\citep{hendrycksmath2021}, using \emph{GPT-4o-mini}~\citep{openai2024gpt4o} as the underlying model for the \textsc{Solver Agent}. We compare it with a CoT-prompted \emph{GPT-4o-mini} baseline for comparison. 

As shown in Table~\ref{tab:performance_comparison}, the TIR-enhanced model consistently outperforms the baseline across all categories. The most significant gains are observed in \textit{Intermediate Algebra} (from 20\% to 36\%) and \textit{Precalculus} (from 18\% to 28\%), where problems demand multi-stage math reasoning and precise computation. Moderate improvements are also observed in \textit{Algebra}, \textit{Number Theory}, and \textit{Prealgebra}, suggesting that even moderately structured problems benefit from our framework. The smallest improvement occurs in \textit{Counting \& Probability}, likely due to the toolkit's limited support for combinatorial logic and discrete structures. Overall, these results confirm the advantage of symbolic tool access with our curated toolkit in enhancing mathematical reasoning capabilities, especially in algebraic domains.

Additionally, we have carried out a qualitative analysis of the failure cases and identified ten distinct types of errors that frequently occur in our TIR framework, spanning issues in both tool usage and symbolic problem solving, shown in \Cref{app:error_analysis}.

\begin{table}[t] 
\caption{Accuracy of GPT-4o-mini on high-difficulty (Level-5) MATH problems across six categories (50 randomly sampled questions for each category).}
\label{tab:performance_comparison} 
\centering 
\begin{tabular}{lcc} 
\toprule 
\textbf{Category} & \textbf{Baseline(\%)}  & \textbf{TIR(\%)} \\
\midrule 
Intermediate Algebra & 20.0 & \textbf{36.0} \\
Precalculus          & 18.0 & \textbf{28.0} \\
Algebra                   & 70.0 & \textbf{80.0} \\
Counting \& Probability   & 50.0 & \textbf{54.0} \\
Number Theory             & 54.0 & \textbf{62.0} \\
Prealgebra                & 58.0 & \textbf{66.0} \\
\bottomrule 
\end{tabular} 
\end{table}

\section{Back-Translating TIR Traces}

While TIR traces offer a structured and powerful approach to mathematical problem solving, they are not directly suitable for training models without access of tools. Our objective is to distill the knowledge from symbolic tools to models, with reasoning traces that are entirely in NL, enabling models to solve math problems without requiring tool access at inference.

\begin{figure*}[t]
    \centering
\includegraphics[width=\textwidth]{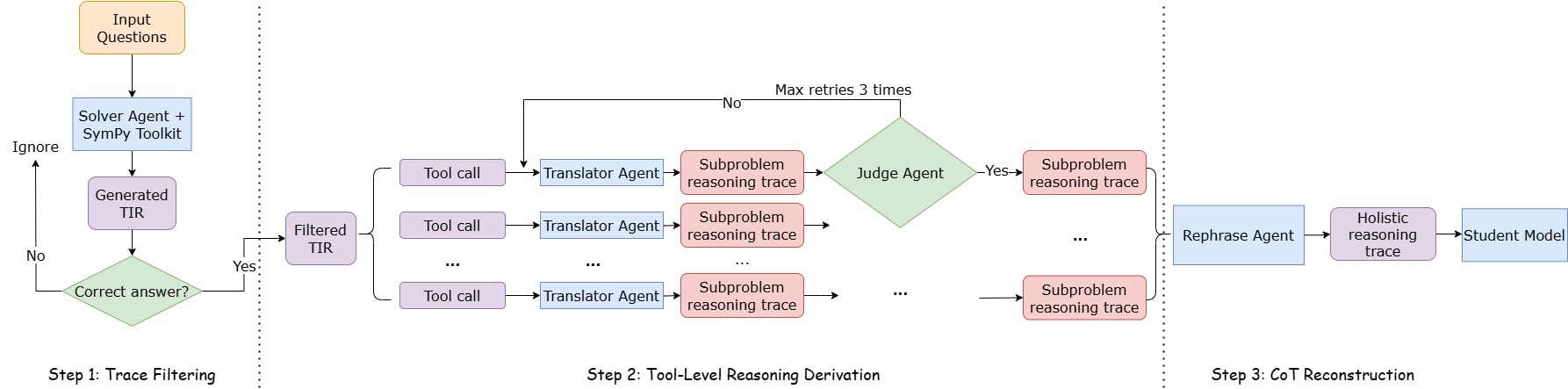}
\vspace{-1em}
    \caption{Overview of the whole pipeline.
    (1) TIR traces are generated by the \textsc{Solver Agent} and are filtered according to their correctness. (2) Each tool call is formulated into a subproblem and a \textsc{Translator Agent} is employed to generate subproblem reasoning trace in NL. A \textsc{Judge Agent} then verifies whether the subproblem trace yields a result equivalent to the original tool call output. (3) Validated subproblem traces are then paired with their corresponding tool calls, and are passed to a \textsc{Rephrase Agent}, which generates a holistic reasoning trace including both high-level planning and detailed reasoning. These traces are used to fine-tune a student model.}
    \label{fig:overallpipeline}
\end{figure*}

As illustrated in Figure~\ref{fig:overallpipeline}, our back-translation pipeline transforms valid TIR traces into NL traces through three key stages: \textit{TIR trace filtering}, \textit{tool-level reasoning derivation}, and \textit{holistic reasoning trace reconstruction}.

\paragraph{TIR Trace Filtering.}  
We begin by filtering all TIR traces generated by the \textsc{Solver Agent} using ground-truth labels provided by math datasets. Only the traces that result in correct final answers are retained for back-translation and the rest which encounter tool call errors or lead to incorrect answers are discarded. This ensures the quality of the reasoning traces forming our training set for distillation.

\paragraph{Tool-Level Reasoning Derivation.}  
Since tool calls in the TIR trace are abstract, typically represented only by the tool name, input, and output, it is difficult to extract meaningful reasoning directly from them. To address this, we transform the full TIR trace into a consistent and interpretable NL trace by generating textual reasoning steps that replace the tool calls. We introduce a \textsc{Translator Agent}: for each tool invocation in the trace, the \textsc{Translator Agent} reformulates the tool call into a subproblem, framed using the tool's name, its docstring (which defines the tool's functionality), and the actual arguments passed. We use the following prompt and let the \textsc{Translator Agent} generate a subproblem reasoning trace that replicates the tool's operation in NL.

\begin{figure}
\begin{tcolorbox}[colframe=black!60, colback=gray!5, sharp corners=south, title=\textsc{Translator Agent} Prompt]
\tcbset{
  compactmdbox/.style={
    enhanced,
    boxrule=0.4pt,
    colback=white,
    colframe=black,
    arc=8mm,              
    boxsep=4pt,
    fontupper=\scriptsize,
    sharp corners=south,  
    width=\columnwidth,
  }
}
\scriptsize
\begin{verbatim}
You are a mathematical problem solver. 
Your task is to solve the following problem
step by step. Clearly illustrate the reasoning 
process and show all calculations explicitly.

Problem: {tool_name}
This problem {docstring}. 
Here is the input required for solving 
this problem: {arguments}

Your final answer should be wrapped in \boxed{}.
Do **not** generate code.  
\end{verbatim}
\end{tcolorbox}

\end{figure}

To ensure the correctness of the generated subproblem trace, a separate \textsc{Judge Agent} is used to compare the \textsc{Translator Agent}'s final answer against the original output of the tool call. Only if both answers are mathematically equivalent do we retain the generated trace. Otherwise, the \textsc{Translator Agent} retries up to three times before we move on to the next tool call. 
Note that the output of each tool call may involve symbolic expressions or rational approximations, depending on the specific computations performed by the tool. However, since the \textsc{Translator Agent} does not have access to the tool's source code, it may be unaware of the expected output format. The introduction of the \textsc{Judge Agent} can help evaluate the reasoning steps produced by the \textsc{Translator Agent} more accurately, reducing false negatives-cases where correct reasoning is penalized solely due to format mismatches with the original tool call output. The prompt for the \textsc{Judge Agent} is shown as follows.

\begin{figure}
\begin{tcolorbox}[colframe=black!70, colback=gray!5, sharp corners=south, title=\textsc{Judge Agent} Prompt]
\scriptsize
\begin{verbatim}
You are a mathematical verifier.
Your task is to verify whether the following 
generated answer is mathematically equivalent 
to the ground truth. 
Ignore the format of the answer. 

Ground Truth: {ground_truth}.

Generated Answer: {final_answer}.

First explain why (or why not) the generated answer 
is mathematically equivalent to the ground truth.
Then clearly state True or False at the end, 
wrapped in \boxed{}.
\end{verbatim}
\end{tcolorbox}

\begin{remark}
\label{remark:remark1}
    If the \textsc{Translator Agent} fails to generate a correct subproblem trace for a tool call after multiple attempts, we retain the original tool call information from the original TIR trace. We avoid discarding them, as each may contain up to 15 tool calls, and strict filtering would result in an unacceptably low acceptance rate. We conducted an empirical investigation over the acceptance rate per tools in \Cref{sec:exp_backtranslation}.
\end{remark}
\end{figure}

We illustrate here a generated subproblem trace based on a tool call from our running example:
\begin{figure}
\begin{tcolorbox}[colframe=black!50, colback=yellow!5, sharp corners=south, boxrule=0.4pt, left=6pt, right=6pt, top=4pt, bottom=4pt]
\scriptsize
$\texttt{solve\_nonlinear\_system}(\{(x - 0)^2 + (y - 0)^2 = 4^2,\ (x - 2)^2 + (y - 0)^2 = 4^2\},\ {[x, y]}) \rightarrow \{(1,\ \sqrt{15}),\ (1,\ -\sqrt{15})\}$
\end{tcolorbox}
\end{figure}
This tool call lacks an explanation of the underlying reasoning. In contrast, a back-translated version derived from our method contains detailed reasoning steps in NL:


\begin{figure}
\begin{tcolorbox}[colframe=brown!80, colback=yellow!5, sharp corners=south, boxrule=0.4pt,
left=6pt, right=6pt, top=4pt, bottom=4pt, before skip=4pt, after skip=4pt, title=Example Output of \textsc{Translator Agent}]
\scriptsize

Let's solve the given system of nonlinear equations step by step.
$
\begin{cases}
(x - 0)^2 + (y - 0)^2 = 4^2 \\
(x - 2)^2 + (y - 0)^2 = 4^2
\end{cases}
$

\textbf{Step 1: Simplify the equations.}
$
x^2 + y^2 = 16, \quad (x - 2)^2 + y^2 = 16
$

\textbf{Step 2: Subtract to eliminate \( y^2 \).}
$
x^2 - (x - 2)^2 = 0 \Rightarrow x^2 - (x^2 - 4x + 4) = 4x - 4 = 0 \Rightarrow x = 1
$

\smallskip\noindent\textbf{Step 3: Substitute \( x = 1 \) into the first equation.}
$
1 + y^2 = 16 \Rightarrow y^2 = 15 \Rightarrow y = \pm \sqrt{15}
$

\smallskip\noindent\textbf{Step 4: Final solution.}
$
\boxed{(x, y) = (1, \sqrt{15}) \quad \text{and} \quad (1, -\sqrt{15})}
$

\end{tcolorbox}
\end{figure}

\paragraph{Holistic Reasoning Trace Reconstruction.}
Since the \textsc{Translator Agent} generates subproblem traces without access to the full TIR trace, inconsistencies may arise when substituting tool calls with them.
To produce self-contained and compact reasoning traces suitable for SFT, we introduce a \textsc{Rephrase Agent} (prompt provided below) that consolidates all intermediate components of a TIR trace, including high-level planning, mathematical explanations, individual tool calls, and their detailed NL-based reasoning into a unified, NL-based reasoning trace.
The final trace removes all tool call information, preserving only the reasoning logic and computation in a fluent narrative that mirrors the structure of the original solution while remaining concise and complete to a student model without tool access at inference.
We include the complete output of \textsc{Rephrase Agent} for our example in \Cref{sec:example_tir_trace}.

\begin{tcolorbox}[colframe=black!60, colback=gray!5, sharp corners=south, title=\textsc{Rephrase Agent} Prompt]
\tcbset{
  compactmdbox/.style={
    enhanced,
    boxrule=0.4pt,
    colback=white,
    colframe=black,
    arc=8mm,              
    boxsep=4pt,
    fontupper=\scriptsize,
    sharp corners=south,  
    width=\columnwidth,
  }
}
\scriptsize
\begin{verbatim}
You are a mathematical answer reformatter. 
Your task is to reformat the given message in 
a clear manner.

You will receive a reasoning path with plans 
and execution steps. The execution steps include 
reasoning based on natural language and tool 
function calls. Remember, do NOT mention using 
SymPy or code in the explanation. 

Your rewritten message should read like a single, 
flowing mathematical solution where each step 
naturally builds upon previous work.
\end{verbatim}
\end{tcolorbox}

\section{Distill Tool Knowledge with SFT over Back-translated Traces}

We finetune using the synthesized reasoning traces obtained through the back-translation pipeline, which combines validated tool-level reasoning with high-level planning. These serve as supervision for training a student model without requiring tool access at inference.

\paragraph{Experimental Setup.} For the base (student) model, we use \emph{Qwen2.5-MATH-7B-Instruct}~\citep{yang2024qwen25math}, which has been instruction-tuned over mathematical domains and serves as a strong baseline for math reasoning tasks. The \textsc{Solver Agent} is based on \emph{GPT-4.1-mini}~\citep{openai2025gpt41}, equipped with the custom-built \texttt{SymPy} toolkit, introduced in \Cref{sec:sympy}. The \textsc{Translator Agent}, \textsc{Judge Agent}, and \textsc{Rephrase Agent} in back-translation are also powered by \emph{GPT-4.1-mini}. Our SFT is performed using Low-Rank Adaptation (LoRA)~\citep{hu2022lora} with a rank of 64, modifying only approximately 1\% of the total parameters. We adopt the \emph{Unsloth}~\citep{unsloth2025} library and BF16 precision for efficient training.
The SFT process is conducted with the following configuration: 3 epochs with a batch size of 1, gradient accumulation steps of 4, learning rate of $2 \times 10^{-4}$, and an AdamW optimizer
with weight decay of 0.01.

\begin{figure*}[t]
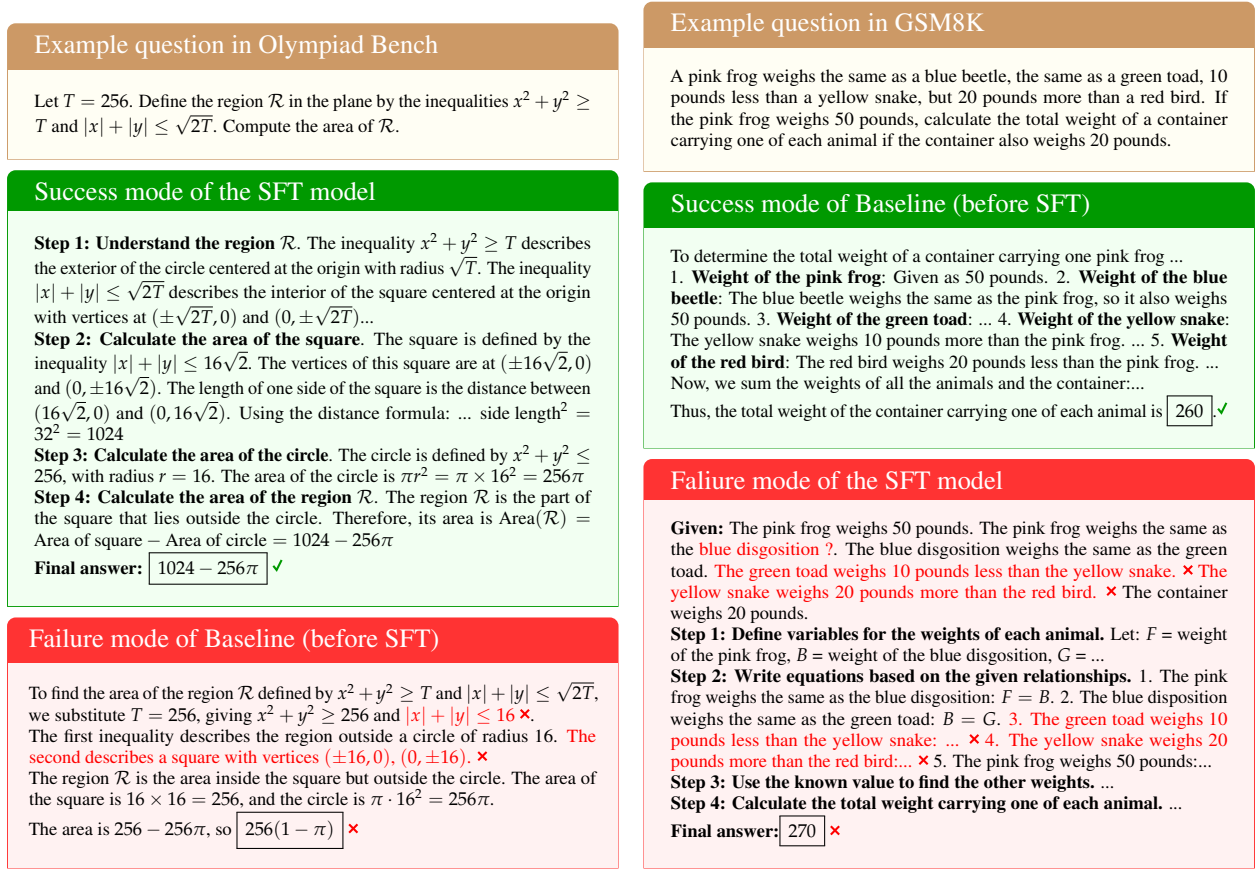

\noindent
\begin{minipage}[b]{0.49\textwidth}
\begin{tcolorbox}[colframe=brown!80, colback=yellow!5, sharp corners=south, boxrule=0.4pt, left=6pt, right=6pt, top=4pt, bottom=4pt, title= Example question in Olympiad Bench]
\scriptsize
Let \( T = 256 \). Define the region \(\mathcal{R}\) in the plane by the inequalities \( x^{2} + y^{2} \geq T \) and \( |x| + |y| \leq \sqrt{2T} \). Compute the area of \(\mathcal{R}\).
\end{tcolorbox}
\begin{tcolorbox}[colframe=green!60!black, colback=green!5, sharp corners=south, boxrule=0.4pt,
left=6pt, right=6pt, top=4pt, bottom=4pt, before skip=4pt, after skip=4pt, title=Success mode of the SFT model]
\scriptsize
\textbf{Step 1: Understand the region \(\mathcal{R}\)}. 
The inequality \( x^{2} + y^{2} \geq T \) describes the exterior of the circle centered at the origin with radius \(\sqrt{T}\). The inequality \( |x| + |y| \leq \sqrt{2T} \) describes the interior of the square centered at the origin with vertices at \((\pm \sqrt{2T}, 0)\) and \((0, \pm \sqrt{2T})\)...

\textbf{Step 2: Calculate the area of the square}.
The square is defined by the inequality \( |x| + |y| \leq 16 \sqrt{2} \). The vertices of this square are at \((\pm 16 \sqrt{2}, 0)\) and \((0, \pm 16 \sqrt{2})\).  
The length of one side of the square is the distance between \((16 \sqrt{2}, 0)\) and \((0, 16 \sqrt{2})\).  
Using the distance formula:  
 ...
$
\text{side length}^2 = 32^2 = 1024
$

\textbf{Step 3: Calculate the area of the circle}.  
The circle is defined by \( x^{2} + y^{2} \leq 256 \), with radius \( r = 16 \). The area of the circle is  
$
\pi r^{2} = \pi \times 16^{2} = 256 \pi
$

\textbf{Step 4: Calculate the area of the region \(\mathcal{R}\)}.  
The region \(\mathcal{R}\) is the part of the square that lies outside the circle. Therefore, its area is  
$
\text{Area}(\mathcal{R}) = \text{Area of square} - \text{Area of circle} = 1024 - 256 \pi
$

\textbf{Final answer:} \(\boxed{1024 - 256 \pi}\) \greentick
\end{tcolorbox}

\begin{tcolorbox}[colframe=red!80, colback=red!5, sharp corners=south, boxrule=0.4pt,
left=4pt, right=4pt, top=3pt, bottom=3pt, before skip=4pt, after skip=4pt, title=Failure mode of Baseline (before SFT)]
\scriptsize
To find the area of the region \(\mathcal{R}\) defined by \(x^2 + y^2 \geq T\) and \(|x| + |y| \leq \sqrt{2T}\), we substitute \(T = 256\), giving \(x^2 + y^2 \geq 256\) and \textcolor{red}{\(|x| + |y| \leq 16\)} \redcross.

The first inequality describes the region outside a circle of radius 16. \textcolor{red}{The second describes a square with vertices \((\pm 16, 0)\), \((0, \pm 16)\).} \redcross

The region \(\mathcal{R}\) is the area inside the square but outside the circle. The area of the square is \(16 \times 16 = 256\), and the circle is \(\pi \cdot 16^2 = 256\pi\).

The area is \(256 - 256\pi\), so \(\boxed{256(1 - \pi)}\) \redcross
\end{tcolorbox}
\end{minipage}
\hfill
\begin{minipage}[b]{0.49\textwidth}
\begin{tcolorbox}[colframe=brown!80, colback=yellow!5, sharp corners=south, boxrule=0.4pt, left=6pt, right=6pt, top=4pt, bottom=4pt,title= Example question in GSM8K]
\scriptsize
A pink frog weighs the same as a blue beetle, the same as a green toad,
10 pounds less than a yellow snake, but 20 pounds more than a red bird.
If the pink frog weighs 50 pounds, calculate the total weight of a container carrying one of each animal if the container also weighs 20 pounds.
\end{tcolorbox}
\begin{tcolorbox}[colframe=green!60!black, colback=green!5, sharp corners=south, boxrule=0.4pt,
left=6pt, right=6pt, top=4pt, bottom=4pt, before skip=4pt, after skip=4pt, title=Success mode of Baseline (before SFT)]
\scriptsize
To determine the total weight of a container carrying one pink frog ...

1. \textbf{Weight of the pink frog}: Given as 50 pounds.
2. \textbf{Weight of the blue beetle}: The blue beetle weighs the same as the pink frog, so it also weighs 50 pounds.
3. \textbf{Weight of the green toad}: ...
4. \textbf{Weight of the yellow snake}: The yellow snake weighs 10 pounds more than the pink frog.
...
5. \textbf{Weight of the red bird}: The red bird weighs 20 pounds less than the pink frog.
...

Now, we sum the weights of all the animals and the container:...

Thus, the total weight of the container carrying one of each animal is \(\boxed{260}\).\greentick

\end{tcolorbox}

\begin{tcolorbox}[colframe=red!80, colback=red!5, sharp corners=south, boxrule=0.4pt,
  left=6pt, right=6pt, top=4pt, bottom=4pt, before skip=4pt, after skip=4pt, title=Faliure mode of the SFT model]
\scriptsize
\textbf{Given:}
The pink frog weighs 50 pounds.
The pink frog weighs the same as the \textcolor{red}{blue disgosition} \textcolor{red}{?}.
The blue disgosition weighs the same as the green toad.
\textcolor{red}{The green toad weighs 10 pounds less than the yellow snake.} \redcross~\textcolor{red}{The yellow snake weighs 20 pounds more than the red bird.} \redcross~The container weighs 20 pounds.

\textbf{Step 1: Define variables for the weights of each animal.}
Let:
\( F \) = weight of the pink frog,
\( B \) = weight of the blue disgosition,
\( G \) = ...

\textbf{Step 2: Write equations based on the given relationships.} 
1. The pink frog weighs the same as the blue disgosition:
$F = B$.
2. The blue disposition weighs the same as the green toad:
$
B = G
$.
\textcolor{red}{3. The green toad weighs 10 pounds less than the yellow snake: ...} \redcross ~\textcolor{red}{4. The yellow snake weighs 20 pounds more than the red bird:...} \redcross ~5. The pink frog weighs 50 pounds:...

\textbf{Step 3: Use the known value to find the other weights.}
...

\textbf{Step 4: Calculate the total weight carrying one of each animal.}
...

\textbf{Final answer:}$\boxed{270}$ \redcross
\end{tcolorbox}

\end{minipage}

\caption{Success and failure cases of SFT model on harder benchmark (Olympiad Bench) and simpler benchmark (GSM8K). See \Cref{box:olympiad-sft-result} for full traces.}
\label{fig: case_study}
\vspace{-1em}
\end{figure*}

\paragraph{Training Data.} 
We generate SFT data based on multiple data sources to ensure diverse and challenging coverage across mathematical domains. 
We first collect the training examples provided by 
\citet{li2025torl}, which consolidates data from several established benchmarks. These include NuminaMATH~\citep{jia2024numinamath}, a curated dataset emphasizing formal symbolic manipulation; MATH~\citep{hendrycksmath2021}, a standard benchmark for high school competition-level mathematics; and DeepScaleR~\citep{deepscaler2025}, a recent dataset designed for step-wise scaling of arithmetic and algebraic difficulty. 
Upon inspection, we observe that the training set contains potential overlap with evaluation data. To mitigate data leakage, we further clean the corpus by filtering out any training examples whose question text shares a repeated 10-gram subsequence with any question in our test sets (following~\citet{deepseek-math}). This deduplication step ensures a fair and reliable assessment of generalization performance. We ended up collecting 25k math problems in total. 
Finally, after the TIR trace filtering process with the \textsc{Solver Agent}, we obtain 11.6k TIR traces, with an overall accuracy around 46\%.

\paragraph{Evaluation.} 
We evaluate model performance using greedy decoding (temperature = 0). We report accuracy on the following benchmarks:
 AIME~\citep{huggingfaceh4_aime_2024}, MATH500~\citep{hendrycksmath2021}, Olympiad Bench~\citep{he2024OlympiadBench}, AMC~\citep{math-ai_amc23}, 
\text{GSM8K}~\citep{cobbe2021gsm8k}. Additionally, we evaluate the final answer using \texttt{math\_verify}~\citep{mathverify}, a robust mathematical expression evaluation system designed to assess the outputs of LLM on mathematical tasks.

\paragraph{Implementation.}
We conduct all experiments using a single NVIDIA H100 80GB GPU for both training and inference. Our agentic pipeline is implemented using the CAMEL framework~\citep{li2023camel}, 
which provides modular support for tool-integrated reasoning and LLM fine-tuning.
We release the synthesized reasoning traces for training at \url{https://huggingface.co/datasets/camel-ai/backtranslated-tir}.

\subsection{Experimental Results}
\begin{table}[t]
    \centering
        \caption{Performance before and after SFT.}
    \label{tab:finetuning_results}
    \renewcommand{\arraystretch}{1.2}
    \begin{tabular}{lcc}
      \toprule
      \textbf{Benchmark} & \multicolumn{1}{c}{\textbf{Baseline (\%)}} & \multicolumn{1}{c}{\textbf{+SFT (\%)}}  \\ 
      \midrule
      AIME             &  7.8 & \textbf{10.0} \\
      Olympiad Bench   & 41.2 & \textbf{43.1  }\\
      AMC              & 51.2 & \textbf{52.4}\\
      MATH500          & \textbf{82.4} & 78.2  \\
      GSM8K            & \textbf{95.5} & 91.1 \\
      \bottomrule
    \end{tabular}

\end{table}
\Cref{tab:finetuning_results} reports the performance of \emph{Qwen2.5-MATH-7B-Instruct} before and after SFT. We find that
SFT on our data leads to clear gains on challenging benchmarks like AIME, AMC, and Olympiad Bench, which require sophisticated multi-step reasoning~\citep{jia2024numinamath}. This also demonstrates that SFT on our generated reasoning traces helps the model internalize the reasoning patterns conveyed by various tool calls, without which it would be unlikely to achieve improved performance on these challenging benchmarks. 
However, slight decline in performance occur on simpler datasets (MATH500, GSM8K), where concise, heuristic-based answers often suffice. 

These trends reflect the influence of our back-translated supervision: the training traces prioritize faithful, step-by-step reasoning, which helps on complex problems but can lead to overly detailed solutions for simpler ones. The model learns to emulate planning followed by symbolic execution, which supports generalization in sophisticated algebraic settings but may reduce performance in tasks that benefit from concise and direct reasoning. To verify this, we do several case studies in \Cref{sec:case study}.

\looseness=-1
\subsection{Case Study}

\label{sec:case study}

We present two case studies in \Cref{fig: case_study} illustrating scenarios where fine-tuning on our reasoning traces improves model performance, and where it leads to degradation.

\looseness=-1
\paragraph{Case 1: Reasoning Decomposition and Tool Knowledge Helps.}
In the Olympiad Bench example, the SFT model decomposes the geometric problem into intermediate subgoals (e.g., identifying similar triangles, computing segment lengths), enabling accurate area calculation. In contrast, the baseline model attempts a direct solution and fails due to incorrect assumptions. This highlights how structured reasoning steps, learned via fine-tuning, are crucial for generalizing to problems that require formal, multi-step mathematical reasoning. 
In addition, we observe that the SFT model learns precise calculations (e.g., in Steps 2 and 3) from tools, indicating that distilling tool knowledge is beneficial for tasks requiring accurate algebraic computation.

\looseness=-1
\paragraph{Case 2: Overcomplication Hurts.}
We show a GSM8K failure where the SFT model is misled by its learned inclination toward exhaustive reasoning. Instead of treating the pink frog as a fixed anchor (50 lbs) and inferring all others relative to it, as the baseline does efficiently, the SFT model introduces unnecessary variables and constraints. Although its derivation is logically sound, this results in a spurious substitution in the reasoning logic (e.g., confusing relative vs. absolute relationships) and ultimately an incorrect final sum (270 v.s. 260). The baseline, on the other hand, applies simpler heuristics and arrives at the correct answer directly.
This highlights a limitation of the model fine-tuned on our generated traces: it may over-reason and overlook simple solutions in cases where brevity or intuitive heuristics are more effective than unnecessarily complicated reasoning.

\looseness=-1
\subsection{Acceptance Rate of \textsc{Translator Agent}}
\label{sec:exp_backtranslation}
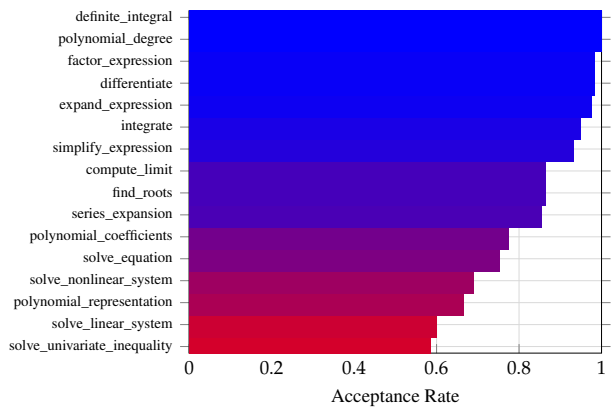
\begin{figure}
    \centering
    \begin{tikzpicture}
    \begin{scope}[scale=0.8]
      \begin{axis}[
        xbar,
        xmin=0, xmax=1.00,
        bar width=12pt,
        enlarge y limits=0.02,
        xlabel={Acceptance Rate},
        xlabel style={font=\small},
        title={},
        title style={font=\normalsize\bfseries, align=center},
        grid=major,
        grid style={line width=.1pt, draw=gray!30},
        tick label style={font=\small},
        ytick={0,1,...,15},
        yticklabels={
          solve\_univariate\_inequality,   
          solve\_linear\_system,           
          polynomial\_representation,      
          solve\_nonlinear\_system,        
          solve\_equation,                 
          polynomial\_coefficients,        
          series\_expansion,               
          find\_roots,                     
          compute\_limit,                  
          simplify\_expression,            
          integrate,                       
          expand\_expression,              
          differentiate,                   
          factor\_expression,              
          polynomial\_degree,              
          definite\_integral               
        },
        yticklabel style={font=\scriptsize, align=right},
        every node near coord/.style={
          font=\tiny,
          /tikz/anchor=west,
          xshift=2pt,
          /pgf/number format/fixed,
          /pgf/number format/precision=4
        },
          point meta=x
      ]
       \addplot+[draw=none,fill=blue!17!red,bar shift=0pt]  coordinates {(0.5845,0)};
  \addplot+[draw=none,fill=blue!20!red,bar shift=0pt]  coordinates {(0.5996,1)};
  \addplot+[draw=none,fill=blue!33!red,bar shift=0pt]  coordinates {(0.6667,2)};
  \addplot+[draw=none,fill=blue!38!red,bar shift=0pt]  coordinates {(0.6902,3)};
  \addplot+[draw=none,fill=blue!51!red,bar shift=0pt]  coordinates {(0.7529,4)};
  \addplot+[draw=none,fill=blue!55!red,bar shift=0pt]  coordinates {(0.7742,5)};
  \addplot+[draw=none,fill=blue!71!red,bar shift=0pt]  coordinates {(0.8542,6)};
  \addplot+[draw=none,fill=blue!73!red,bar shift=0pt]  coordinates {(0.8634,7)};
  \addplot+[draw=none,fill=blue!73!red,bar shift=0pt]  coordinates {(0.8649,8)};
  \addplot+[draw=none,fill=blue!86!red,bar shift=0pt]  coordinates {(0.9313,9)};
  \addplot+[draw=none,fill=blue!90!red,bar shift=0pt]  coordinates {(0.9492,10)};
  \addplot+[draw=none,fill=blue!95!red,bar shift=0pt]  coordinates {(0.9750,11)};
  \addplot+[draw=none,fill=blue!97!red,bar shift=0pt]  coordinates {(0.9840,12)};
  \addplot+[draw=none,fill=blue!97!red,bar shift=0pt]  coordinates {(0.9842,13)};
  \addplot+[draw=none,fill=blue!100!red,bar shift=0pt] coordinates {(1.0000,14)};
  \addplot+[draw=none,fill=blue!100!red,bar shift=0pt] coordinates {(1.0000,15)};
      \end{axis}
      \end{scope}
    \end{tikzpicture}
    \vspace{-1em}
    \caption{Acceptance rate for each tool.}
    \label{fig:pgfplots_tool_accuracy_sorted}
    \vspace{-1em}
  \end{figure}

As mentioned in \Cref{remark:remark1}, we provide here a detailed analysis over the rejection rate of \textsc{Translator Agent} across the tool calls belonging to different tools.

\Cref{fig:pgfplots_tool_accuracy_sorted} reports the acceptance rate for the generated subproblem traces of each tool, sorted from lowest to highest. 
We observe a stratification in performance: tools involving algebraic calculation and equation solving (e.g., \texttt{solve\_univariate\_inequality}, \texttt{solve\_}\texttt{linear\_system}) exhibit lower acceptance rate, reflecting the complexity and ambiguity in translating their intermediate reasoning into pure language. In contrast, tools with more surface-level computation, such as \texttt{polynomial\_degree}, \texttt{factor\_expression}, and \texttt{definite\_integral}, achieve perfect or near-perfect acceptance rate. 
This trend points to a promising direction: while some tool behaviors require more advanced language modeling for faithful back-translation, others can already be captured with high fidelity, enabling small models to acquire symbolic knowledge through language supervision alone. The quality of back-translation also depends on the capabilities of the \textsc{Translator Agent}, as more powerful models generate more accurate and complete reasoning traces, especially for tools involves more complicated reasoning, highlighting the importance of improving the reasoning quality of the translation model itself.

\looseness=-1
\section{Conclusion}
We present a framework for enhancing mathematical reasoning in language models by converting TIR traces into back-translated NL, enabling fine-tuning without tool access at inference. This back-translation pipeline allows models to learn from tool-augmented solutions while producing purely NL outputs. Empirical results show that small open-source models fine-tuned on these traces benefit from both embedded tool knowledge and structured solution patterns, achieving performance gains on competition-level math benchmarks. We further analyze the success and failure modes of the fine-tuned model, identifying key factors that contribute to its improvement or degradation.

\textbf{Limitations.} Our method heavily depends on the capabilities of the \textsc{Translator Agent}, \textsc{Judge Agent}, and \textsc{Rephrase Agent}, which may fail on edge cases or ambiguous tool outputs. The filtering strategy is intentionally strict, retaining only fully correct traces, which can limit training diversity and introduce biases from the \textsc{Solver Agent}. Additionally, our current toolkit does not cover all mathematical domains (such as geometry or discrete combinatorics) and our experiments are restricted to a single student model. Notably, while our synthesized traces improve performance on complex tasks, they can degrade performance on simpler math datasets, likely due to over-reasoning or unnecessary complexity. Addressing these limitations is key to achieving broader applicability and generalization.

\section*{Impact Statement}

This work introduces a method for enhancing mathematical reasoning in language models by fine-tuning on NL traces distilled from symbolic tool calls. Our approach enables smaller models to learn precise and interpretable reasoning patterns without requiring tool execution at inference time. The resulting models have potential applications in education, scientific computing, and formal verification. We release our code and data to support reproducibility and future research. No human data is involved, and we do not foresee significant risks of misuse.

\section*{Acknowledgment}

This work was conducted as a collaborative open-source research initiative at \url{http://CAMEL-AI.org}, supported by funding from Eigent.AI. We also thank the Qwen teams for their generous support on computational resources.

\bibliographystyle{unsrtnat}
\bibliography{main,rlhf} 

\clearpage
\appendix
\onecolumn

\section{Qualitative Analysis of Failure Cases}  
\label{app:error_analysis}

\begin{figure*}
  \centering
  \begin{tikzpicture}
  \begin{scope}
    \pie[
      /tikz/every pin/.style={align=center,font=\footnotesize},
      font=\scriptsize,
      rotate=160,
      radius=2,
      pos={8,0}, explode=0.1,
      hide number
    ]{
      10.29/Misuse of Tool (10.29\%),
       7.35/Error in Tool Usage (7.35\%),
      22.06/Did Not Use Tool (22.06\%),
       1.47/Did Not Follow Tool (1.47\%),
       8.82/Lack of Tool (8.82\%),
      11.76/Tool Limitation (11.76\%),
             4.41/Precision Loss (4.41\%),
      11.76/Overcomplicated Calculation (11.76\%),
      22.06/Mistake in Reasoning (22.06\%),
       1.47/Evaluation Error (1.47\%)
    }
      \end{scope}
  \end{tikzpicture}

  \caption{Distribution of error types observed in Level-5 problems.}
  \label{fig:enter-label}
\end{figure*}
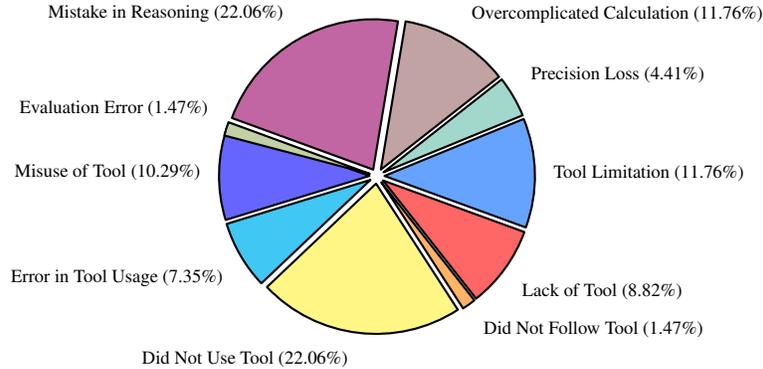

Despite clear improvements from tool integration, performance on the most challenging categories-\textit{Intermediate Algebra} and \textit{Precalculus}-remains limited. To better understand the limitations of our approach, we conduct a detailed error analysis on these two domains. Each category includes 50 Level-5 problems, totaling 100 questions. Of these, 68 were answered incorrectly by the TIR-enhanced model.
 We categorize these failures into ten types across two broad groups: \textit{errors involving tool usage} and \textit{errors without tool usage}. The distribution of these errors is shown in Figure~\ref{fig:enter-label}. This error distribution highlights key challenges in both tool invocation and logical reasoning, which we analyze in more detail below.

\subsection{Errors Involving Tool Use}

\paragraph{Misuse of Tool.} 
The model occasionally selects an inappropriate tool despite the availability of one that would enable successful problem solving. For instance, in a problem requiring constrained optimization using the method of Lagrange multipliers, the model instead attempts to apply a general-purpose equation solver. This misalignment between tool capability and problem structure results in failure, even though the task is solvable with the provided toolkit.

\paragraph{Error in Tool Usage.} These errors occur when the correct tool is selected but misused due to syntax errors or input formatting issues. For instance, the model may express absolute values as \texttt{|x|} rather than the syntactically valid \texttt{Abs(x)}. In another case, it inputs angles in degrees despite the tool expecting radians, resulting in invalid outputs.

\paragraph{Did Not Use Tool.} In problems requiring symbolic manipulation or arithmetic simplification, the model occasionally attempts a fully manual solution despite having access to a suitable tool. These attempts often fail due to algebraic mistakes or overlooked simplifications.

\paragraph{Did Not Follow Tool Output.} Sometimes, the model invokes a tool correctly and receives a valid output, but disregards it in the subsequent reasoning. For example, in a pre-calculus question, an answer is given by a tool to compute $\sin(6*x) + \cos(4*x) = 0$, but is disregarded or misunderstood, leading to a manual computation and thus an incorrect final solution.

\paragraph{Lack of Tool.}  
Some problems require operations outside the scope of the current toolkit, forcing the model to attempt manual solutions or rely on poor approximations. For instance, the toolkit lacks support for \textbf{geometry-related operations} such as vector projections and angle computations, which are hard to replicate symbolically. It also lacks native support for \textbf{combinatorial reasoning}, such as computing binomial coefficients or counting arrangements. In another case, the model repeatedly tries to evaluate $\sin(175/2)$ but becomes stuck, as the tool only returns symbolic forms rather than usable numerical values. These cases highlight the limitations of fixed-function toolkits in handling the full range of mathematical reasoning required at high difficulty levels.

\paragraph{Precision Loss.}  
When performing symbolic computations involving complex numbers or floating-point arithmetic, the model may correctly execute a tool call but fail to interpret or round the result appropriately. For instance, in a problem involving simplification of a complex expression, the model receives a symbolic output from the tool such as  
\[
5.99955009573917 - 0.0882302939421837i.
\]  
Mathematically, the problem requires only the real part to be considered-perhaps because the imaginary part is negligible, spurious, or canceled by context. However, instead of rounding the real part to the correct final answer \( \boxed{6} \), the model either outputs the full complex number or approximates the real part as \( 5.999 \), resulting in a mismatch against the expected exact answer.

\paragraph{Tool Limitation.}  
Even with correct tool selection and input formatting, the model may fail due to inherent solver constraints. For example, when solving the cubic equation \( 9x^3 - 20x = 8\sqrt{2} \), the \texttt{SymPy} tool is unable to return an analytical real root, leaving the model unable to complete the solution. Such problems could be handled by more advanced systems like WolframAlpha~\citep{wolframalpha}, but the current toolkit lacks sufficient symbolic depth. These cases highlight the need for expanded solver capabilities and fallback mechanisms when standard routines are inadequate.

\subsection{Errors Without Tool Use}

\paragraph{Overcomplicated Calculation.}  
The model sometimes adopts overly complex strategies when simpler methods would suffice, increasing the chance of errors. For example, in a problem requiring analysis of inequality constraints, the model attempts symbolic optimization via differentiation, despite the solution being obtainable through direct simplification. Such over-engineering introduces unnecessary tool calls and complicates otherwise straightforward reasoning.

\paragraph{Mistake in Reasoning.}  
Logical errors in the model's solution paths are common, even when individual steps appear valid. For instance, it incorrectly reasons about angular constraints, asserting that the interval $135^\circ < x < 180^\circ$ corresponds to $\frac{3\pi}{4} < x < \pi$, then selecting $x = \frac{5\pi}{4}$-a value clearly outside the valid domain. Elsewhere, the model misinterprets the goal of a geometric problem: when asked to determine when no triangle can be formed, it instead solves for conditions under which triangles \emph{can} be formed. These failures underscore a combination of weak symbolic planning, misunderstanding of mathematical constraints, and insufficient awareness of tool behavior when documentation is limited.

\paragraph{Evaluation Error.} Even after reaching the correct answer, the model may fail to present it in the expected format. For instance, it might omit enclosing the final answer in \verb|\boxed{}|, causing automatic evaluation scripts to mark it incorrect.

\subsection{Toward More Robust TIR}  
To address the common failure modes observed, we propose several directions for improving tool-integrated mathematical reasoning systems:

\begin{enumerate}
    \item \textbf{Tool Awareness via Documentation and Signatures.}  
    Models often misuse tools due to incomplete understanding of their behavior. Providing structured tool documentation, function signatures, and example usages-either via system prompts or training corpora-can help models align tool selection with problem requirements.

    \item \textbf{Planning.}  
    Errors in reasoning often arise from premature or inappropriate tool calls, especially for complex questions. Incorporating \textbf{planning} modules that decide \emph{when} and \emph{how} to invoke tools can lead to more coherent reasoning traces. 

    \item \textbf{Interactive and Reflective Tool Use.}  
    Effective mathematical reasoning requires tools to be used iteratively-not in a single forward pass. Models should learn to call a tool, interpret its output, reflect on its relevance, and adapt their next step accordingly. This loop of \emph{tool use $\rightarrow$ reflection $\rightarrow$ follow-up} mirrors expert human problem solving and is critical for handling multi-stage dependencies and ambiguous subgoals.

    \item \textbf{Unified Representations for Tool I/O.}  
    Inconsistent input/output handling (e.g., misinterpreting radians vs degrees) contributes to tool misuse. Standardizing representations and enforcing validation layers around tool interfaces could minimize such semantic mismatches.

    \item \textbf{Expanding Tool Coverage.}  
    Existing toolkits primarily support algebraic and calculus-based operations, limiting their applicability across diverse mathematical domains. To improve generality, future systems should integrate tools that support a wider range of reasoning tasks, including geometric, combinatorial, and structural problem types, enabling more comprehensive symbolic capabilities.

\end{enumerate}

\section{System Prompt for \textsc{Solver Agent}}
\label{app:system_prompt}

\vspace{1em}
\begin{tcolorbox}[colframe=black!70, colback=gray!5, sharp corners=south, title=System Prompt for \textsc{Solver Agent}, fonttitle=\bfseries]
\scriptsize
\begin{verbatim}
You are a helpful assistant that solves math problems using the given tools. 

You should reflect on your execution result, refine your plan, and output the next action. 

Not all occasions need to use tools; it depends on the nature of the problem.

IMPORTANT: When using mathematical expressions, use SymPy grammar. For example:
  - Use ** for exponents (e.g., x**2 for x²)
  - Use sqrt() for square roots
  - Use pi for Pi
  - Use * for multiplication (e.g., 2*x not 2x, x*y not xy)
  - Use solve() for solving equations
  - Use Abs() for absolute values, do not use |.| (e.g., Abs(-4) not |-4|)
  - Use Sum (capitalized) to represent summation over a series (e.g., Sum(ceil(log(i, 2)), (i, 2, 1000)))
  - Use capital I for imaginary numbers
  - Each variable must be a single letter (e.g., replace lambda with l)
  - Always convert angles to radians, not degrees
  - Use . between vectors as inner product (e.g., (u + v) . (2*u - v))
  - Use Matrix to represent coordinates, not tuples
  - All string input must conform to sympy.parse format
  - For coordinates, do not write P(x,y,z); instead, use (x, y, z)

  For equations:
    * NEVER use = in expressions
    * Use Eq(left_side, right_side) (e.g., Eq(y, (-80 - 320*i)/x))
    * Alternatively, rearrange to expression = 0 form

If you obtain the final answer, provide it in LaTeX wrapped in \boxed{{}}. 
Only reply in LaTeX and analytic solution.

**NEVER** write the code block explicitly.
\end{verbatim}
\end{tcolorbox}

\section{Example TIR Trace and Back-translated Trace}
\label{sec:example_tir_trace}
\begin{tcolorbox}[colframe=brown!80, colback=yellow!5, sharp corners=south, boxrule=0.4pt, left=6pt, right=6pt, top=4pt, bottom=4pt,title=Question]
\small
Let ( $C_1$ ) and ( $C_2$ ) be circles of radius 1 that are in the same plane and tangent to each other. Determine how many circles of radius 3 can be placed in this plane such that they are tangent to both ( $C_1$ ) and ( $C_2$ ).
\end{tcolorbox}

\begin{tcolorbox}[title=Tool-Integrated Reasoning Trace,
  colframe=brown!80, colback=yellow!5,
  fonttitle=\bfseries,
  sharp corners=south,
  enhanced,
  breakable]
  \scriptsize
\begin{verbatim}
<message>
Step-by-step plan:

1. **Understand the problem setup**: We have two circles \( C_1 \) and \( C_2 \) of radius 1, 
tangent to each other. We want to find how many circles of radius 3 can be tangent to both \( C_1 \) 
and \( C_2 \).

2. **Set up coordinate system and variables**: Place \( C_1 \) and \( C_2 \) in the coordinate 
plane for convenience. For example, place \( C_1 \) at the origin and \( C_2 \) at (2,0) since they 
are tangent and each has radius 1.

3. **Express conditions for tangency**:
- The circle of radius 3 must be tangent to both \( C_1 \) and \( C_2 \).
- Use the distance between centers condition: the distance between the center of the new circle and 
the center of \( C_1 \) must be either 3 + 1 = 4 (external tangency) or |3 - 1| = 2 (internal tangency).
- Similarly for \( C_2 \).

4. **Formulate equations**: Let the center of the new circle be \((x,y)\). Write equations for the 
distances from \((x,y)\) to the centers of \( C_1 \) and \( C_2 \) equal to 4 or 2, depending on tangency type.

5. **Solve the system of equations**: Use algebraic or geometric methods to find all possible 
\((x,y)\) that satisfy the tangency conditions.

6. **Count the number of solutions**: The number of valid \((x,y)\) solutions corresponds to the 
number of circles of radius 3 tangent to both \( C_1 \) and \( C_2 \).

Tools to use:

- **Coordinate geometry**: To set up the problem in a coordinate plane and express conditions as equations.
- **Distance formula**: To express tangency conditions as equations involving distances between centers.
- **Equation solving (algebraic)**: To solve the system of equations for \((x,y)\).
- **Geometric reasoning**: To interpret the solutions and count the number of valid circles.

I expect to find a finite number of solutions for \((x,y)\), each corresponding to a circle of 
radius 3 tangent to both \( C_1 \) and \( C_2 \).
</message>

<message>
Step 1: Understand the problem setup.

We have two circles \( C_1 \) and \( C_2 \), each with radius 1, tangent to each other. We want to 
find how many circles of radius 3 can be tangent to both \( C_1 \) and \( C_2 \).

Next, I will proceed to Step 2: Set up a coordinate system and variables. I will place \( C_1 \) 
and \( C_2 \) in the coordinate plane for convenience, placing \( C_1 \) at the origin (0,0) and \( 
C_2 \) at (2,0) since they are tangent and each has radius 1.

I will use coordinate geometry to set up the problem in the coordinate plane.
Let's place the centers of the two circles \( C_1 \) and \( C_2 \) as follows:

- Center of \( C_1 \) at \( (0,0) \) with radius 1.
- Center of \( C_2 \) at \( (2,0) \) with radius 1.

Next, I will move to Step 3: Express the conditions for tangency of the new circle of radius 3 to 
both \( C_1 \) and \( C_2 \).

I will use the distance formula and the tangency conditions to write equations for the center 
\((x,y)\) of the new circle.
</message>

<tool>
<tool_name>solve_nonlinear_system</tool_name>
<args>{"sympy_equations": ["(x-0)**2 + (y-0)**2 - 4**2", "(x-2)**2 + (y-0)**2 - 4**2"], 
"variables": ["x", "y"]}</args>
<result>Tuple(Integer(1), Pow(Integer(15), Rational(1, 2))), Tuple(Integer(1), Mul(Integer(-1), 
Pow(Integer(15), Rational(1, 2))))</result>
</tool>

<tool>
<tool_name>solve_nonlinear_system</tool_name>
<args>{"sympy_equations": ["(x-0)**2 + (y-0)**2 - 4**2", "(x-2)**2 + (y-0)**2 - 2**2"], 
"variables": ["x", "y"]}</args>
<result>Tuple(Integer(4), Integer(0))</result>
</tool>

<tool>
<tool_name>solve_nonlinear_system</tool_name>
<args>{"sympy_equations": ["(x-0)**2 + (y-0)**2 - 2**2", "(x-2)**2 + (y-0)**2 - 4**2"], 
"variables": ["x", "y"]}</args>
<result>Tuple(Integer(-2), Integer(0))</result>
</tool>

<tool>
<tool_name>solve_nonlinear_system</tool_name>
<args>{"sympy_equations": ["(x-0)**2 + (y-0)**2 - 2**2", "(x-2)**2 + (y-0)**2 - 2**2"], 
"variables": ["x", "y"]}</args>
<result>Tuple(Integer(1), Pow(Integer(3), Rational(1, 2))), Tuple(Integer(1), Mul(Integer(-1), 
Pow(Integer(3), Rational(1, 2))))</result>
</tool>

<message>
We consider the possible tangency distances between the new circle of radius 3 and the two circles \( C_1 \) 
and \( C_2 \) of radius 1. The distance between centers for tangency can be either 4 (external tangency) or 
2 (internal tangency).

We solve the system of equations for the center \((x,y)\) of the new circle under the following cases:

1. Both distances are 4:
\[
(x-0)^2 + (y-0)^2 = 4^2 = 16
\]
\[
(x-2)^2 + (y-0)^2 = 4^2 = 16
\]
Solutions: \((1, \sqrt{15})\) and \((1, -\sqrt{15})\).

2. Distance to \( C_1 \) is 4, distance to \( C_2 \) is 2:
\[
(x-0)^2 + (y-0)^2 = 16
\]
\[
(x-2)^2 + (y-0)^2 = 4
\]
Solution: \((4,0)\).

3. Distance to \( C_1 \) is 2, distance to \( C_2 \) is 4:
\[
(x-0)^2 + (y-0)^2 = 4
\]
\[
(x-2)^2 + (y-0)^2 = 16
\]
Solution: \((-2,0)\).

4. Both distances are 2:
\[
(x-0)^2 + (y-0)^2 = 4
\]
\[
(x-2)^2 + (y-0)^2 = 4
\]
Solutions: \((1, \sqrt{3})\) and \((1, -\sqrt{3})\).

Thus, the possible centers of the new circle are:
\[
(1, \sqrt{15}), (1, -\sqrt{15}), (4,0), (-2,0), (1, \sqrt{3}), (1, -\sqrt{3})
\]

Each corresponds to a circle of radius 3 tangent to both \( C_1 \) and \( C_2 \).

Therefore, the number of such circles is 6.

\[
\boxed{6}
\]
</message>


\end{verbatim}
\end{tcolorbox}

\begin{tcolorbox}[title=Backtranslated Chain-of-Thoughts per Tool Call,
  colframe=brown!80, colback=yellow!5,
  fonttitle=\bfseries,
  sharp corners=south,
  enhanced,
  breakable]
  \scriptsize
\
\begin{verbatim}
<tool>
<tool_name>solve_nonlinear_system</tool_name>
<args>{""sympy_equations"": [""(x-0)**2 + (y-0)**2 - 4**2"", ""(x-2)**2 + (y-0)**2 - 4**2""], 
""variables"": [""x"", ""y""]}</args>
<cot>
Chain of Thought for solve_nonlinear_system:
Let's solve the given system of nonlinear equations step by step.
### Given:
Equations:
1. ((x - 0)^2 + (y - 0)^2 - 4^2 = 0)
2. ((x - 2)^2 + (y - 0)^2 - 4^2 = 0)
Variables: (x, y)
---
### Step 1: Rewrite the equations clearly
1. (x^2 + y^2 = 16) (since (4^2 = 16))
2. ((x - 2)^2 + y^2 = 16)
---
### Step 2: Analyze the system
These are equations of two circles:
- Circle 1: Center at ((0,0)), radius (4)
- Circle 2: Center at ((2,0)), radius (4)
We want to find the points ((x,y)) where these two circles intersect.
---
### Step 3: Subtract the two equations to eliminate (y^2)
\[
x^2 + y^2 = 16
\]
\[
(x - 2)^2 + y^2 = 16
\]
Subtract the second from the first:
\[
x^2 + y^2 - \left[(x - 2)^2 + y^2\right] = 16 - 16
\]
\[
x^2 - (x - 2)^2 = 0
\]
Expand ((x - 2)^2):
\[
x^2 - (x^2 - 4x + 4) = 0
\]
\[
x^2 - x^2 + 4x - 4 = 0
\]
\[
4x - 4 = 0
\]
\[
4x = 4
\]
\[
x = 1
\]
---
### Step 4: Substitute (x = 1) back into one of the original equations to find (y)
Using the first equation:
\[
x^2 + y^2 = 16
\]
\[
1^2 + y^2 = 16
\]
\[
1 + y^2 = 16
\]
\[
y^2 = 15
\]
\[
y = \pm \sqrt{15}
\]
---
### Step 5: Write the solutions
The solutions are:
\[
(x, y) = (1, \sqrt{15}) \quad \text{and} \quad (1, -\sqrt{15})
\]
---
### Final answer:
\[
\boxed{
\left\[
(1, \sqrt{15}), \quad (1, -\sqrt{15})
\right\]
}
\]
</cot>
</tool>
<tool>
<tool_name>solve_nonlinear_system</tool_name>
<args>{""sympy_equations"": [""(x-0)**2 + (y-0)**2 - 4**2"", ""(x-2)**2 + (y-0)**2 - 2**2""], 
""variables"": [""x"", ""y""]}</args>
<cot>
Chain of Thought for solve_nonlinear_system:
Let's solve the given system of nonlinear equations step by step.
### Given:
Equations:
1. ((x - 0)^2 + (y - 0)^2 - 4^2 = 0)
2. ((x - 2)^2 + (y - 0)^2 - 2^2 = 0)
Variables: (x, y)
---
### Step 1: Rewrite the equations clearly
1. (x^2 + y^2 = 16) (since (4^2 = 16))
2. ((x - 2)^2 + y^2 = 4) (since (2^2 = 4))
---
### Step 2: Express the system
\[
\begin{cases}
x^2 + y^2 = 16 \\
(x - 2)^2 + y^2 = 4
\end{cases}
\]
---
### Step 3: Subtract the second equation from the first to eliminate (y^2)
\[
(x^2 + y^2) - ((x - 2)^2 + y^2) = 16 - 4
\]
Simplify:
\[
x^2 - (x - 2)^2 = 12
\]
Expand ((x - 2)^2):
\[
x^2 - (x^2 - 4x + 4) = 12
\]
Simplify inside the parentheses:
\[
x^2 - x^2 + 4x - 4 = 12
\]
This reduces to:
\[
4x - 4 = 12
\]
---
### Step 4: Solve for (x)
\[
4x = 12 + 4 = 16
\]
\[
x = \frac{16}{4} = 4
\]
---
### Step 5: Substitute (x = 4) into one of the original equations to find (y)
Use the first equation:
\[
x^2 + y^2 = 16
\]
Substitute (x = 4):
\[
4^2 + y^2 = 16
\]
\[
16 + y^2 = 16
\]
\[
y^2 = 0
\]
\[
y = 0
\]
---
### Step 6: Verify the solution in the second equation
\[
(x - 2)^2 + y^2 = 4
\]
Substitute (x = 4, y = 0):
\[
(4 - 2)^2 + 0^2 = 4
\]
\[
2^2 = 4
\]
\[
4 = 4
\]
True, so the solution is valid.
---
### Step 7: Check if there are other solutions
The system represents two circles:
- Circle 1: center ((0,0)), radius 4
- Circle 2: center ((2,0)), radius 2
The circles may intersect at more than one point.
Let's find the other intersection points.
---
### Step 8: From Step 3, we had:
\[
4x - 4 = 12 \implies 4x = 16 \implies x = 4
\]
This gives one intersection point at (x=4).
But let's check if there are other points by solving the system more generally.
---
### Alternative approach:
From the two equations:
\[
x^2 + y^2 = 16 \quad (1)
\]
\[
(x - 2)^2 + y^2 = 4 \quad (2)
\]
Subtract (2) from (1):
\[
x^2 + y^2 - (x - 2)^2 - y^2 = 16 - 4
\]
\[
x^2 - (x^2 - 4x + 4) = 12
\]
\[
x^2 - x^2 + 4x - 4 = 12
\]
\[
4x = 16 \implies x = 4
\]
So the only possible (x) value for intersection is 4.
Substitute (x=4) into (1):
\[
16 + y^2 = 16 \implies y^2 = 0 \implies y=0
\]
So the only intersection point is ((4,0)).
---
### Step 9: Check if the circles are tangent or intersect at one point
Since the distance between centers is:

\[
d = \sqrt{(2-0)^2 + (0-0)^2} = 2
\]
Sum of radii:
\[
r_1 + r_2 = 4 + 2 = 6
\]
Difference of radii:
\[
|r_1 - r_2| = |4 - 2| = 2
\]
Since (d = |r_1 - r_2|), the circles are tangent internally, touching at exactly one point.
---
### Final solution:
\[
\boxed{(4, 0)}
\]
---
### Expressing the solution as required:
The solution set is (\{(4, 0)\}).
---
[
\boxed{
\text{JSON string: } \{""result"": [[4, 0]]\}
}
]
</cot>
</tool>
<tool>
<tool_name>solve_nonlinear_system</tool_name>
<args>{""sympy_equations"": [""(x-0)**2 + (y-0)**2 - 2**2"", ""(x-2)**2 + (y-0)**2 - 4**2""], 
""variables"": [""x"", ""y""]}</args>
<cot>
Chain of Thought for solve_nonlinear_system:
Let's solve the given system of nonlinear equations step by step.
### Given:
Equations:
1. ((x - 0)^2 + (y - 0)^2 - 2^2 = 0)
2. ((x - 2)^2 + (y - 0)^2 - 4^2 = 0)
Variables: (x, y)
---
### Step 1: Rewrite the equations clearly
1. (x^2 + y^2 = 4) (since (2^2 = 4))
2. ((x - 2)^2 + y^2 = 16) (since (4^2 = 16))
---
### Step 2: Express the system explicitly
\[
\begin{cases}
x^2 + y^2 = 4 \\
(x - 2)^2 + y^2 = 16
\end{cases}
\]
---
### Step 3: Subtract the first equation from the second to eliminate (y^2)
\[
(x - 2)^2 + y^2 - (x^2 + y^2) = 16 - 4
\]
Simplify:
\[
(x - 2)^2 - x^2 = 12
\]
Expand ((x - 2)^2):
\[
(x^2 - 4x + 4) - x^2 = 12
\]
Simplify:
\[
-4x + 4 = 12
\]
---
### Step 4: Solve for (x)
\[
-4x = 12 - 4 = 8
\]
\[
x = -2
\]
---
### Step 5: Substitute (x = -2) into the first equation to find (y)
\[
(-2)^2 + y^2 = 4
\]
\[
4 + y^2 = 4
\]
\[
y^2 = 0
\]
\[
y = 0
\]
---
### Step 6: Verify the solution in the second equation
\[
(x - 2)^2 + y^2 = 16
\]
\[
(-2 - 2)^2 + 0^2 = 16
\]
\[
(-4)^2 = 16
\]
\[
16 = 16
\]
True.
---
### Step 7: Check if there are other solutions
The subtraction step gave a unique (x), but let's check if the system might have more solutions.
Alternatively, solve for (y^2) from the first equation:
\[
y^2 = 4 - x^2
\]
Substitute into the second equation:
\[
(x - 2)^2 + (4 - x^2) = 16
\]
Expand:
\[
x^2 - 4x + 4 + 4 - x^2 = 16
\]
Simplify:
\[
-4x + 8 = 16
\]
\[
-4x = 8
\]
\[
x = -2
\]
So again, (x = -2).
Then (y^2 = 4 - (-2)^2 = 4 - 4 = 0), so (y = 0).
No other solutions.
---
### Final solution:
\[
(x, y) = (-2, 0)
\]
---
### Expressing the solution as requested:
The solution tuple corresponds to ((x, y)):
\[
\boxed{[(-2, 0)]}
\]
---
### Summary:
- The system represents two circles:
- Circle 1: center at ((0,0)), radius 2
- Circle 2: center at ((2,0)), radius 4
- The circles intersect at exactly one point: ((-2, 0))
---
(\boxed{[(-2, 0)]})
</cot>
</tool>
<tool>
<tool_name>solve_nonlinear_system</tool_name>
<args>{""sympy_equations"": [""(x-0)**2 + (y-0)**2 - 2**2"", ""(x-2)**2 + (y-0)**2 - 2**2""], 
""variables"": [""x"", ""y""]}</args>
<cot>
Chain of Thought for solve_nonlinear_system:
Let's solve the given system of nonlinear equations step by step.
### Given:
Equations:
1. ((x - 0)^2 + (y - 0)^2 - 2^2 = 0)
2. ((x - 2)^2 + (y - 0)^2 - 2^2 = 0)
Variables: (x, y)
---
### Step 1: Rewrite the equations clearly
1. (x^2 + y^2 = 4) (since (2^2 = 4))
2. ((x - 2)^2 + y^2 = 4)
---
### Step 2: Analyze the system
- The first equation represents a circle centered at ((0,0)) with radius 2.
- The second equation represents a circle centered at ((2,0)) with radius 2.
We want to find the points ((x,y)) where these two circles intersect.
---
### Step 3: Subtract the two equations to eliminate (y^2)
\[
(x - 2)^2 + y^2 = 4
\]
\[
x^2 + y^2 = 4
\]
Subtract the second from the first:
\[
(x - 2)^2 + y^2 - (x^2 + y^2) = 4 - 4
\]
\[
(x - 2)^2 - x^2 = 0
\]
Expand ((x - 2)^2):
\[
(x^2 - 4x + 4) - x^2 = 0
\]
\[
-4x + 4 = 0
\]
---
### Step 4: Solve for (x)
\[
-4x + 4 = 0 \implies -4x = -4 \implies x = 1
\]
---
### Step 5: Substitute (x = 1) into one of the original equations to find (y)
Use the first equation:
\[
x^2 + y^2 = 4
\]
\[
1^2 + y^2 = 4
\]
\[
1 + y^2 = 4
\]
\[
y^2 = 3
\]
\[
y = \pm \sqrt{3}
\]
---
### Step 6: Write the solutions
The solutions are:
\[
(x, y) = (1, \sqrt{3}) \quad \text{and} \quad (1, -\sqrt{3})
\]
---
### Final answer:
\[
\boxed{
\left\{ (1, \sqrt{3}), (1, -\sqrt{3}) \right\}
}
\]
</cot>
</tool>

\end{verbatim}
\end{tcolorbox}

\begin{tcolorbox}[title= Holistic Chain-of-Thought obtained after \textsc{Rephrase Agent},
  colframe=brown!80, colback=yellow!5,
  fonttitle=\bfseries,
  sharp corners=south,
  enhanced,
  breakable
]
\scriptsize
\begin{verbatim}
""**Problem:**
Let \( C_1 \) and \( C_2 \) be two circles of radius 1 in the same plane, tangent to each other. 
Determine how many circles of radius 3 can be placed in this plane such that each is tangent to 
both \( C_1 \) and \( C_2 \).

---

### Step 1: Understand the problem setup

We have two circles \( C_1 \) and \( C_2 \), each with radius 1, tangent to each other. We want to 
find all circles of radius 3 that are tangent to both \( C_1 \) and \( C_2 \).

---

### Step 2: Set up a coordinate system

Place the two given circles conveniently in the coordinate plane:

- Let the center of \( C_1 \) be at the origin \( (0,0) \).
- Since \( C_1 \) and \( C_2 \) are tangent and each has radius 1, place the center of \( C_2 \) at 
\( (2,0) \).

Thus:

- \( C_1: \) center \( O_1 = (0,0) \), radius \( r_1 = 1 \).
- \( C_2: \) center \( O_2 = (2,0) \), radius \( r_2 = 1 \).

---

### Step 3: Conditions for tangency of the new circle

Let the new circle have radius \( R = 3 \) and center \( (x,y) \).

For the new circle to be tangent to \( C_1 \), the distance between centers must be either:

- \( R + r_1 = 3 + 1 = 4 \) (external tangency), or
- \( |R - r_1| = 3 - 1 = 2 \) (internal tangency).

Similarly, for tangency to \( C_2 \), the distance between centers must be either 4 or 2.

Therefore, the center \( (x,y) \) of the new circle must satisfy:

\[
\text{Distance}( (x,y), O_1 ) = d_1 \in \{2,4\}
\]
\[
\text{Distance}( (x,y), O_2 ) = d_2 \in \{2,4\}
\]

---

### Step 4: Formulate and solve the systems of equations

We consider all possible combinations of \( (d_1, d_2) \) from the set \(\{2,4\} \times \{2,4\}\):

---

#### Case 1: \( d_1 = 4 \), \( d_2 = 4 \)

\[
(x - 0)^2 + (y - 0)^2 = 16
\]
\[
(x - 2)^2 + (y - 0)^2 = 16
\]

Subtracting the two equations to eliminate \( y^2 \):

\[
x^2 + y^2 = 16
\]
\[
(x - 2)^2 + y^2 = 16
\]

Subtract:

\[
x^2 - (x - 2)^2 = 0
\]

Expand:

\[
x^2 - (x^2 - 4x + 4) = 0 \implies 4x - 4 = 0 \implies x = 1
\]

Substitute \( x=1 \) into the first equation:

\[
1^2 + y^2 = 16 \implies y^2 = 15 \implies y = \pm \sqrt{15}
\]

**Solutions:**

\[
(1, \sqrt{15}), \quad (1, -\sqrt{15})
\]

---

#### Case 2: \( d_1 = 4 \), \( d_2 = 2 \)

\[
x^2 + y^2 = 16
\]
\[
(x - 2)^2 + y^2 = 4
\]

Subtract:

\[
x^2 - (x - 2)^2 = 12
\]

Expand:

\[
x^2 - (x^2 - 4x + 4) = 12 \implies 4x - 4 = 12 \implies 4x = 16 \implies x = 4
\]

Substitute \( x=4 \) into the first equation:

\[
16 + y^2 = 16 \implies y^2 = 0 \implies y = 0
\]

**Solution:**

\[
(4, 0)
\]

---

#### Case 3: \( d_1 = 2 \), \( d_2 = 4 \)

\[
x^2 + y^2 = 4
\]
\[
(x - 2)^2 + y^2 = 16
\]

Subtract:

\[
(x - 2)^2 - x^2 = 12
\]

Expand:

\[
(x^2 - 4x + 4) - x^2 = 12 \implies -4x + 4 = 12 \implies -4x = 8 \implies x = -2
\]

Substitute \( x = -2 \) into the first equation:

\[
4 + y^2 = 4 \implies y^2 = 0 \implies y = 0
\]

**Solution:**

\[
(-2, 0)
\]

---

#### Case 4: \( d_1 = 2 \), \( d_2 = 2 \)

\[
x^2 + y^2 = 4
\]
\[
(x - 2)^2 + y^2 = 4
\]

Subtract:

\[
(x - 2)^2 - x^2 = 0
\]

Expand:

\[
x^2 - 4x + 4 - x^2 = 0 \implies -4x + 4 = 0 \implies x = 1
\]

Substitute \( x=1 \) into the first equation:

\[
1 + y^2 = 4 \implies y^2 = 3 \implies y = \pm \sqrt{3}
\]

**Solutions:**

\[
(1, \sqrt{3}), \quad (1, -\sqrt{3})
\]

---

### Step 5: Collect all solutions

The centers of the circles of radius 3 tangent to both \( C_1 \) and \( C_2 \) are:

\[
(1, \sqrt{15}), \quad (1, -\sqrt{15}), \quad (4, 0), \quad (-2, 0), \quad (1, \sqrt{3}), 
\quad (1, -\sqrt{3})
\]

Each corresponds to a distinct circle of radius 3 tangent to both \( C_1 \) and \( C_2 \).

---

### **Final answer:**

\[
\boxed{6}
\]

There are **6** such circles of radius 3 tangent to both given circles of radius 1.""
\end{verbatim}
\end{tcolorbox}

We also present another example question and back-translated traces below:

\begin{tcolorbox}[colframe=brown!80, colback=yellow!5, sharp corners=south, boxrule=0.4pt, left=6pt, right=6pt, top=4pt, bottom=4pt,title=Question 2]
\small
Calculate the remainder of the Euclidean division of $2022^{2023^{2024}}$ by 19.
\end{tcolorbox}

\begin{tcolorbox}[title=Holistic Chain-of-Thought obtained after \texttt{Rephrase Agent} 2,
  colframe=brown!80, colback=yellow!5,
  fonttitle=\bfseries,
  sharp corners=south,
  enhanced,
  breakable]
  \scriptsize
\begin{verbatim}
""We want to find the remainder when \(2022^{2023^{2024}}\) is divided by 19, i.e., compute
\[
2022^{2023^{2024}} \mod 19.
\]

---

### Step 1: Reduce the base modulo 19

Since the remainder depends only on the base modulo 19, we first compute:
\[
2022 \mod 19.
\]

Divide 2022 by 19:
\[
2022 \div 19 = 106 \text{ remainder } 8,
\]
because \(19 \times 106 = 2014\) and \(2022 - 2014 = 8\).

Thus,
\[
2022 \equiv 8 \pmod{19}.
\]

So the expression simplifies to:
\[
8^{2023^{2024}} \mod 19.
\]

---

### Step 2: Use Fermat's Little Theorem to reduce the exponent modulo 18

Since 19 is prime, Fermat's Little Theorem states that for any integer \(a\) not divisible by 19,
\[
a^{18} \equiv 1 \pmod{19}.
\]

Therefore, the exponent can be reduced modulo 18:
\[
8^{2023^{2024}} \equiv 8^{(2023^{2024} \mod 18)} \pmod{19}.
\]

We need to compute:
\[
e = 2023^{2024} \mod 18.
\]

---

### Step 3: Reduce the base of the exponent modulo 18

Calculate:
\[
2023 \mod 18.
\]

Divide 2023 by 18:
\[
2023 \div 18 = 112 \text{ remainder } 7,
\]
since \(18 \times 112 = 2016\) and \(2023 - 2016 = 7\).

Thus,
\[
2023 \equiv 7 \pmod{18}.
\]

So,
\[
e = 7^{2024} \mod 18.
\]

---

### Step 4: Compute \(7^{2024} \mod 18\)

To simplify this, we use Euler's theorem. First, compute Euler's totient \(\varphi(18)\).

Since
\[
18 = 2 \times 3^2,
\]
we have
\[
\varphi(18) = \varphi(2) \times \varphi(3^2) = 1 \times (9 - 3) = 6.
\]

Because \(\gcd(7, 18) = 1\), Euler's theorem applies:
\[
7^6 \equiv 1 \pmod{18}.
\]

Reduce the exponent modulo 6:
\[
2024 \mod 6 = 2024 - 6 \times 337 = 2024 - 2022 = 2.
\]

Therefore,
\[
7^{2024} \equiv 7^2 \pmod{18}.
\]

Calculate \(7^2 \mod 18\):
\[
7^2 = 49,
\]
and
\[
49 \mod 18 = 49 - 2 \times 18 = 49 - 36 = 13.
\]

Hence,
\[
7^{2024} \equiv 13 \pmod{18}.
\]

---

### Step 5: Calculate the final remainder

We now have:
\[
8^{2023^{2024}} \equiv 8^{13} \pmod{19}.
\]

Compute \(8^{13} \mod 19\).

Calculate powers of 8 modulo 19 step-by-step:

- \(8^1 \equiv 8 \pmod{19}\)
- \(8^2 = 64 \equiv 64 - 3 \times 19 = 64 - 57 = 7 \pmod{19}\)
- \(8^3 = 8^2 \times 8 = 7 \times 8 = 56 \equiv 56 - 2 \times 19 = 56 - 38 = 18 \pmod{19}\)
- \(8^4 = 8^3 \times 8 = 18 \times 8 = 144 \equiv 144 - 7 \times 19 = 144 - 133 = 11 \pmod{19}\)
- \(8^5 = 8^4 \times 8 = 11 \times 8 = 88 \equiv 88 - 4 \times 19 = 88 - 76 = 12 \pmod{19}\)
- \(8^6 = 8^5 \times 8 = 12 \times 8 = 96 \equiv 96 - 5 \times 19 = 96 - 95 = 1 \pmod{19}\)

Since \(8^6 \equiv 1 \pmod{19}\), powers of 8 repeat every 6 steps modulo 19.

Express 13 as:
\[
13 = 6 \times 2 + 1.
\]

Therefore,
\[
8^{13} = 8^{6 \times 2 + 1} = (8^6)^2 \times 8^1 \equiv 1^2 \times 8 = 8 \pmod{19}.
\]

---

### Final answer:

\[
\boxed{8}.
\]

The remainder when \(2022^{2023^{2024}}\) is divided by 19 is **8**.""

\end{verbatim}
\end{tcolorbox}

\section{\texttt{SymPy} Toolkit Definition}
\label{app:sympy_toolkit}
Here we include the full definition of our custom-defined \texttt{SymPy} toolkit. 

\begin{tcolorbox}[title=\texttt{SymPy} Toolkit Definition,
  colback=gray!5,
  colframe=blue!50!black,
  coltitle=white,
  colbacktitle=blue!30!black,
  fonttitle=\bfseries,
  sharp corners=south,
  enhanced,
  breakable]
  \scriptsize
    \begin{verbatim}
import json
from typing import List, Optional

def simplify_expression(self, expression: str) -> str:
    r"""Simplifies a mathematical expression.

    Args:
        expression (str): The mathematical expression to simplify,
            provided as a string.

    Returns:
        str: JSON string containing the simplified mathematical
            expression in the `"result"` field. If an error occurs,
            the `"status"` field will be set to `"error"` with a
            corresponding `"message"`.
    """
    import sympy as sp

    try:
        expr = sp.parsing.sympy_parser.parse_expr(expression)
        simplified = sp.simplify(expr)
        return json.dumps(
            {"status": "success", "result": str(simplified)},
            ensure_ascii=False,
        )
    except Exception as e:
        return self.handle_exception("simplify_expression", e)

def expand_expression(self, expression: str) -> str:
    r"""Expands an algebraic expression.

    Args:
        expression (str): The algebraic expression to expand,
            provided as a string.

    Returns:
        str: JSON string containing the expanded algebraic expression
            in the `"result"` field. If an error occurs, the JSON
            string will include an `"error"` field with the corresponding
            error message.
    """
    import sympy as sp

    try:
        expr = sp.parsing.sympy_parser.parse_expr(expression)
        expanded_expr = sp.expand(expr)
        return json.dumps(
            {"result": str(expanded_expr)}, ensure_ascii=False
        )
    except Exception as e:
        return self.handle_exception("expand_expression", e)

def factor_expression(self, expression: str) -> str:
    r"""Factors an algebraic expression.

    Args:
        expression (str): The algebraic expression to factor,
            provided as a string.

    Returns:
        str: JSON string containing the factored algebraic expression
            in the `"result"` field. If an error occurs, the JSON string
            will include an `"error"` field with the corresponding error
            message.
    """
    import sympy as sp

    try:
        expr = sp.parsing.sympy_parser.parse_expr(expression)
        factored_expr = sp.factor(expr)
        return json.dumps(
            {"result": str(factored_expr)}, ensure_ascii=False
        )
    except Exception as e:
        return self.handle_exception("factor_expression", e)

def solve_linear_system(
    self, equations: List[str], variables: List[str]
) -> str:
    r"""Solves a system of linear equations.

    Args:
        equations (List[str]): A list of strings representing the linear
            equations to be solved.
        variables (List[str]): A list of strings representing the variables
            involved in the equations.

    Returns:
        str: JSON string containing the solution to the system of equations
            in the `"result"` field. Each solution is represented as
            a tuple of values corresponding to the variables. If an
            error occurs, the JSON string will include an `"error"`
            field with the corresponding error message.
    """
    import sympy as sp

    try:
        eqs = [sp.sympify(eq) for eq in equations]
        vars = sp.symbols(variables)
        solution = sp.linsolve(eqs, vars)
        return json.dumps(
            {"result": [str(sol) for sol in solution]}, ensure_ascii=False
        )
    except Exception as e:
        return self.handle_exception("solve_linear_system", e)

def solve_nonlinear_system(
    self, sympy_equations: List[str], variables: List[str]
) -> str:
    r"""Solves a system of nonlinear equations.

    Args:
        sympy_equations (List[str]): A list of strings representing the
            nonlinear equations to be solved. The equation to solve, must
            be compatible with SymPy, provided as a string.

        variables (List[str]): A list of strings representing the variables
            involved in the equations.

    Returns:
        str: JSON string containing the solutions to the system of
            equations in the `"result"` field. Each solution is
            represented as a tuple of values corresponding to the
            variables. If an error occurs, the JSON string will
            include an `"error"` field with the corresponding
            error message.
    """
    import sympy as sp

    try:
        eqs = [sp.sympify(eq) for eq in sympy_equations]
        vars = sp.symbols(variables)
        solution = sp.nonlinsolve(eqs, vars)
        return json.dumps(
            {"result": [str(sol) for sol in solution]}, ensure_ascii=False
        )
    except Exception as e:
        return self.handle_exception("solve_nonlinear_system", e)

def solve_univariate_inequality(
    self, inequality: str, variable: str
) -> str:
    r"""Solves a single-variable inequality.

    Args:
        inequality (str): A string representing the inequality
            to be solved.
        variable (str): The variable in the inequality.

    Returns:
        str: JSON string containing the solution to the inequality in the
            `"result"` field. The solution is represented in a symbolic
            format (e.g., intervals or expressions). If an error occurs,
            the JSON string will include an `"error"` field with the
            corresponding error message.
    """
    import sympy as sp

    try:
        var = sp.symbols(variable)
        ineq = sp.sympify(inequality)
        solution = sp.solve_univariate_inequality(ineq, var)
        return json.dumps({"result": str(solution)}, ensure_ascii=False)
    except Exception as e:
        return self.handle_exception("solve_univariate_inequality", e)

def reduce_inequalities(self, inequalities: List[str]) -> str:
    r"""Reduces a system of inequalities.

    Args:
        inequalities (List[str]): A list of strings representing the
            inequalities to be reduced.

    Returns:
        str: JSON string containing the reduced system of inequalities
            in the `"result"` field. The solution is represented in
            a symbolic format (e.g., combined intervals or expressions).
            If an error occurs, the JSON string will include an `"error"`
            field with the corresponding error message.
    """
    import sympy as sp

    try:
        ineqs = [sp.sympify(ineq) for ineq in inequalities]
        solution = sp.reduce_inequalities(ineqs)
        return json.dumps({"result": str(solution)}, ensure_ascii=False)
    except Exception as e:
        return self.handle_exception("reduce_inequalities", e)

def polynomial_representation(self, expression: str, variable: str) -> str:
    r"""Represents an expression as a polynomial.

    Args:
        expression (str): The mathematical expression to represent as
            a polynomial, provided as a string.
        variable (str): The variable with respect to which the polynomial
            representation will be created.

    Returns:
        str: JSON string containing the polynomial representation of the
            expression in the `"result"` field. The polynomial is returned
            in a symbolic format. If an error occurs, the JSON string will
            include an `"error"` field with the corresponding error
            message.
    """

    import sympy as sp

    try:
        var = sp.symbols(variable)
        expr = sp.parsing.sympy_parser.parse_expr(expression)
        poly = sp.Poly(expr, var)
        return json.dumps({"result": str(poly)}, ensure_ascii=False)
    except Exception as e:
        return self.handle_exception("polynomial_representation", e)

def polynomial_degree(self, expression: str, variable: str) -> str:
    r"""Returns the degree of a polynomial.

    Args:
        expression (str): The polynomial expression for which the degree
            is to be determined, provided as a string.
        variable (str): The variable with respect to which the degree
            of the polynomial is calculated.

    Returns:
        str: JSON string containing the degree of the polynomial in the
            `"result"` field. If an error occurs, the JSON string will
            include an `"error"` field with the corresponding error
            message.
    """
    import sympy as sp

    try:
        var = sp.symbols(variable)
        expr = sp.parsing.sympy_parser.parse_expr(expression)
        degree = int(sp.degree(expr, var))
        return json.dumps({"result": degree}, ensure_ascii=False)
    except Exception as e:
        return self.handle_exception("polynomial_degree", e)

def polynomial_coefficients(self, expression: str, variable: str) -> str:
    r"""Returns the coefficients of a polynomial.

    Args:
        expression (str): The polynomial expression from which the
            coefficients are to be extracted, provided as a string.
        variable (str): The variable with respect to which the polynomial
            coefficients are determined.

    Returns:
        str: JSON string containing the list of coefficients of the
            polynomial in the `"result"` field. The coefficients are
            ordered from the highest degree term to the constant term.
            If an error occurs, the JSON string will include an `"error"
            field with the corresponding error message.
    """
    import sympy as sp

    try:
        var = sp.symbols(variable)
        expr = sp.parsing.sympy_parser.parse_expr(expression)
        coeffs = sp.Poly(expr, var).all_coeffs()
        return json.dumps(
            {"result": [str(coeff) for coeff in coeffs]},
            ensure_ascii=False,
        )
    except Exception as e:
        return self.handle_exception("polynomial_coefficients", e)

def solve_equation(
    self, sympy_equation: str, variable: Optional[str] = None
) -> str:
    r"""Solves an equation for a specific variable.

    Args:
        sympy_equation(str): The equation to solve, must be compatible
            with SymPy, provided as a string.
        variable (str, optional): The variable to solve for. If not
            specified, the function will use the default variable.

    Returns:
        str: JSON string containing the solutions to the equation in the
            `"result"` field. Each solution is represented as a string.
            If an error occurs, the JSON string will include an `"error"`
            field with the corresponding error message.
    """
    import sympy as sp

    try:
        variable = (
            sp.symbols(variable)
            if variable
            else sp.symbols(self.default_variable)
        )
        eq = sp.sympify(sympy_equation)
        solutions = sp.solve(eq, variable)
        return json.dumps(
            {"result": [str(sol) for sol in solutions]}, ensure_ascii=False
        )
    except Exception as e:
        return self.handle_exception("solve_equation", e)

def find_roots(self, expression: str) -> str:
    r"""Finds the roots of a polynomial or algebraic equation.

    Args:
        expression (str): The polynomial or algebraic equation for which
            the roots are to be found, provided as a string.

    Returns:
        str: JSON string containing the roots of the expression in the
            `"result"` field. The roots are represented as a list of
            solutions. If an error occurs, the JSON string will include
            a `"status"` field set to `"error"` and a `"message"` field
            with the corresponding error description.
    """
    import sympy as sp

    try:
        expr = sp.parsing.sympy_parser.parse_expr(expression)
        roots = sp.solve(expr)
        return json.dumps(
            {"status": "success", "result": str(roots)}, ensure_ascii=False
        )

    except Exception as e:
        return self.handle_exception("find_roots", e)

def differentiate(
    self, expression: str, variable: Optional[str] = None
) -> str:
    r"""Differentiates an expression with respect to a variable.

    Args:
        expression (str): The mathematical expression to differentiate,
            provided as a string.
        variable (str, optional): The variable with respect to which the
            differentiation is performed. If not specified, the default
            variable is used.

    Returns:
        str: JSON string containing the derivative of the expression in the
            `"result"` field. If an error occurs, the JSON string will
            include an `"error"` field with the corresponding error
            message.
    """
    import sympy as sp

    try:
        variable = (
            sp.symbols(variable)
            if variable
            else sp.symbols(self.default_variable)
        )
        expr = sp.parsing.sympy_parser.parse_expr(expression)
        derivative = sp.diff(expr, variable)
        return json.dumps({"result": str(derivative)}, ensure_ascii=False)
    except Exception as e:
        return self.handle_exception("differentiate", e)

def integrate(
    self, expression: str, variable: Optional[str] = None
) -> str:
    r"""Integrates an expression with respect to a variable.

    Args:
        expression (str): The mathematical expression to integrate,
            provided as a string.
        variable (str, optional): The variable with respect to which the
            integration is performed. If not specified, the default
            variable is used.

    Returns:
        str: JSON string containing the integral of the expression in the
            `"result"` field. If an error occurs, the JSON string will
            include an `"error"` field with the corresponding error
            message.
    """
    import sympy as sp

    try:
        variable = (
            sp.symbols(variable)
            if variable
            else sp.symbols(self.default_variable)
        )
        expr = sp.parsing.sympy_parser.parse_expr(expression)
        integral = sp.integrate(expr, variable)
        return json.dumps({"result": str(integral)}, ensure_ascii=False)
    except Exception as e:
        return self.handle_exception("integrate", e)

def definite_integral(
    self, expression: str, variable: str, lower: float, upper: float
) -> str:
    r"""Computes the definite integral of an expression within given
    bounds.

    Args:
        expression (str): The mathematical expression to integrate,
            provided as a string.
        variable (str): The variable with respect to which the definite
            integration is performed.
        lower (float): The lower limit of the integration.
        upper (float): The upper limit of the integration.

    Returns:
        str: JSON string containing the result of the definite integral
            in the `"result"` field. If an error occurs, the JSON string
            will include an `"error"` field with the corresponding error
            message.
    """
    import sympy as sp

    try:
        var = sp.symbols(variable)
        expr = sp.parsing.sympy_parser.parse_expr(expression)
        integral = sp.integrate(expr, (var, lower, upper))
        return json.dumps({"result": str(integral)}, ensure_ascii=False)
    except Exception as e:
        return self.handle_exception("definite_integral", e)

def series_expansion(
    self, expression: str, variable: str, point: float, order: int
) -> str:
    r"""Expands an expression into a Taylor series around a given point up
    to a specified order.

    Args:
        expression (str): The mathematical expression to expand, provided
            as a string.
        variable (str): The variable with respect to which the series
            expansion is performed.
        point (float): The point around which the Taylor series is
            expanded.
        order (int): The order up to which the series expansion is
        computed.

    Returns:
        str: JSON string containing the Taylor series expansion of the
            expression in the `"result"` field. If an error occurs,
            the JSON string will include an `"error"` field with the
            corresponding error message.
    """
    import sympy as sp

    try:
        var = sp.symbols(variable)
        expr = sp.parsing.sympy_parser.parse_expr(expression)
        series = sp.series(expr, var, point, order)
        return json.dumps({"result": str(series)}, ensure_ascii=False)
    except Exception as e:
        return self.handle_exception("series_expansion", e)

def compute_limit(
    self,
    expression: str,
    variable: str,
    point: float,
) -> str:
    r"""Computes the limit of an expression as a variable approaches
    a point.

    Args:
        expression (str): The mathematical expression for which the limit
            is to be computed, provided as a string.
        variable (str): The variable with respect to which the limit is
            computed.
        point (float): The point that the variable approaches.

    Returns:
        str: JSON string containing the computed limit of the expression
            in the `"result"` field. If an error occurs, the JSON string
            will include an `"error"` field with the corresponding error
            message.
    """
    import sympy as sp

    try:
        var = sp.symbols(variable)
        expr = sp.parsing.sympy_parser.parse_expr(expression)
        limit = sp.limit(expr, var, point)
        return json.dumps({"result": str(limit)}, ensure_ascii=False)
    except Exception as e:
        return self.handle_exception("compute_limit", e)

def find_critical_points(self, expression: str, variable: str) -> str:
    r"""Finds the critical points of an expression by setting its
    derivative to zero.

    Args:
        expression (str): The mathematical expression for which critical
            points are to be found, provided as a string.
        variable (str): The variable with respect to which the critical
            points are determined.

    Returns:
        str: JSON string containing the critical points of the expression
            in the `"result"` field. The critical points are returned as a
            list of values corresponding to the variable. If an error
            occurs, the JSON string will include an `"error"` field with
            the corresponding error message.
    """
    import sympy as sp

    try:
        var = sp.symbols(variable)
        expr = sp.parsing.sympy_parser.parse_expr(expression)
        derivative = sp.diff(expr, var)
        critical_points = sp.solve(derivative, var)
        return json.dumps(
            {"result": [str(point) for point in critical_points]},
            ensure_ascii=False,
        )
    except Exception as e:
        return self.handle_exception("find_critical_points", e)

def check_continuity(
    self, expression: str, variable: str, point: float
) -> str:
    r"""Checks if an expression is continuous at a given point.

    Args:
        expression (str): The mathematical expression to check for
            continuity, provided as a string.
        variable (str): The variable with respect to which continuity
            is checked.
        point (float): The point at which the continuity of the expression
            is checked.

    Returns:
        str: JSON string containing the result of the continuity check in
            the `"result"` field. The result will be `"True"` if the
            expression is continuous at the given point, otherwise
            `"False"`. If an error occurs, the JSON string will include
            an `"error"` field with the corresponding error message.
    """
    import sympy as sp

    try:
        var = sp.symbols(variable)
        expr = sp.parsing.sympy_parser.parse_expr(expression)
        left_limit = sp.limit(expr, var, point, dir='-')
        right_limit = sp.limit(expr, var, point, dir='+')
        value_at_point = expr.subs(var, point)
        is_continuous = left_limit == right_limit == value_at_point
        return json.dumps(
            {"result": str(is_continuous)}, ensure_ascii=False
        )
    except Exception as e:
        return self.handle_exception("check_continuity", e)

def compute_determinant(self, matrix: List[List[float]]) -> str:
    r"""Computes the determinant of a matrix.

    Args:
        matrix (List[List[float]]): A two-dimensional list representing
            the matrix for which the determinant is to be computed.

    Returns:
        str: JSON string containing the determinant of the matrix in the
            `"result"` field. If an error occurs, the JSON string will
            include an `"error"` field with the corresponding error
            message.
    """
    import sympy as sp

    try:
        mat = sp.Matrix(matrix)
        determinant = mat.det()
        return json.dumps({"result": str(determinant)}, ensure_ascii=False)
    except Exception as e:
        return self.handle_exception("compute_determinant", e)

def compute_inverse(self, matrix: List[List[float]]) -> str:
    r"""Computes the inverse of a matrix.

    Args:
        matrix (List[List[float]]): A two-dimensional list representing
            the matrix for which the inverse is to be computed.

    Returns:
        str: JSON string containing the inverse of the matrix in the
            `"result"` field. The inverse is represented in a symbolic
            matrix format. If an error occurs, the JSON string will
            include an `"error"` field with the corresponding error
            message.
    """
    import sympy as sp

    try:
        mat = sp.Matrix(matrix)
        inverse = mat.inv()
        return json.dumps({"result": str(inverse)}, ensure_ascii=False)
    except Exception as e:
        return self.handle_exception("compute_inverse", e)

def compute_eigenvalues(self, matrix: List[List[float]]) -> str:
    r"""Computes the eigenvalues of a matrix.

    Args:
        matrix (List[List[float]]): A two-dimensional list representing
            the matrix for which the eigenvalues are to be computed.

    Returns:
        str: JSON string containing the eigenvalues of the matrix in the
            `"result"` field. The eigenvalues are represented as a
            dictionary where keys are the eigenvalues (as strings) and
            values are their multiplicities (as strings). If an error
            occurs, the JSON string will include an `"error"` field
            with the corresponding error message.
    """
    import sympy as sp

    try:
        mat = sp.Matrix(matrix)
        eigenvalues = mat.eigenvals()
        return json.dumps(
            {"result": {str(k): str(v) for k, v in eigenvalues.items()}},
            ensure_ascii=False,
        )
    except Exception as e:
        return self.handle_exception("compute_eigenvalues", e)

def compute_eigenvectors(self, matrix: List[List[float]]) -> str:
    r"""Computes the eigenvectors of a matrix.

    Args:
        matrix (List[List[float]]): A two-dimensional list representing
            the matrix for which the eigenvectors are to be computed.

    Returns:
        str: JSON string containing the eigenvectors of the matrix in the
            `"result"` field. Each eigenvalue is represented as a
            dictionary with the following keys:
            - `"eigenvalue"`: The eigenvalue (as a string).
            - `"multiplicity"`: The multiplicity of the eigenvalue
            (as an integer).
            - `"eigenvectors"`: A list of eigenvectors
            (each represented as a string).

            If an error occurs, the JSON string will include an `"error"`
            field with the corresponding error message.
    """
    import sympy as sp

    try:
        mat = sp.Matrix(matrix)
        eigenvectors = mat.eigenvects()
        result = [
            {
                "eigenvalue": str(eigenvalue),
                "multiplicity": multiplicity,
                "eigenvectors": [str(v) for v in vectors],
            }
            for eigenvalue, multiplicity, vectors in eigenvectors
        ]
        return json.dumps({"result": result}, ensure_ascii=False)
    except Exception as e:
        return self.handle_exception("compute_eigenvectors", e)

def compute_nullspace(self, matrix: List[List[float]]) -> str:
    r"""Computes the null space of a matrix.

    Args:
        matrix (List[List[float]]): A two-dimensional list representing
            the matrix for which the null space is to be computed.

    Returns:
        str: JSON string containing the null space of the matrix in the
            `"result"` field. The null space is represented as a list of
            basis vectors, where each vector is given as a string in
            symbolic format. If an error occurs, the JSON string will
            include an `"error"` field with the corresponding error
            message.
    """
    import sympy as sp

    try:
        mat = sp.Matrix(matrix)
        nullspace = mat.nullspace()
        return json.dumps(
            {"result": [str(vec) for vec in nullspace]}, ensure_ascii=False
        )
    except Exception as e:
        return self.handle_exception("compute_nullspace", e)

def compute_rank(self, matrix: List[List[float]]) -> str:
    r"""Computes the rank of a matrix.

    Args:
        matrix (List[List[float]]): A two-dimensional list representing
            the matrix for which the rank is to be computed.

    Returns:
        str: JSON string containing the rank of the matrix in the
            `"result"` field. The rank is represented as an integer.
            If an error occurs,the JSON string will include an
            `"error"` field with the corresponding error message.
    """
    import sympy as sp

    try:
        mat = sp.Matrix(matrix)
        rank = mat.rank()
        return json.dumps({"result": rank}, ensure_ascii=False)
    except Exception as e:
        return self.handle_exception("compute_rank", e)

def compute_inner_product(
    self, vector1: List[float], vector2: List[float]
) -> str:
    r"""Computes the inner (dot) product of two vectors.

    Args:
        vector1 (List[float]): The first vector as a list of floats.
        vector2 (List[float]): The second vector as a list of floats.

    Returns:
        str: JSON string containing the inner product in the `"result"`
            field. If an error occurs, the JSON string will include an
            `"error"` field with the corresponding error message.

    Raises:
        ValueError: If the vectors have different dimensions.
    """
    import sympy as sp

    try:
        # Convert the lists into sympy Matrix objects (column vectors)
        v1 = sp.Matrix(vector1)
        v2 = sp.Matrix(vector2)

        # Check that the vectors have the same dimensions.
        if v1.shape != v2.shape:
            raise ValueError(
                "Vectors must have the same dimensions to compute "
                "the inner product."
            )

        # Compute the dot (inner) product.
        inner_product = v1.dot(v2)
        return json.dumps(
            {"result": str(inner_product)}, ensure_ascii=False
        )
    except Exception as e:
        return self.handle_exception("compute_inner_product", e)

def handle_exception(self, func_name: str, error: Exception) -> str:
    r"""Handles exceptions by logging and returning error details.

    Args:
        func_name (str): The name of the function where the
        exception occurred.
        error (Exception): The exception object containing
        details about the error.

    Returns:
        str: JSON string containing the error details.
            The JSON includes:
            - `"status"`: Always set to `"error"`.
            - `"message"`: A string representation of the
            exception message.
    """
    logger.error(f"Error in {func_name}: {error}")
    return json.dumps(
        {"status": "error", "message": f"Error in {func_name}: {error}"},
        ensure_ascii=False,
    )

    
    \end{verbatim}
\end{tcolorbox}
\section{Full traces in \Cref{sec:case study}}
\begin{tcolorbox}[title=Olympiad Bench Sample Problem Definition,
                                     colback=yellow!5,
                                     colframe=brown!80,
                                     fonttitle=\bfseries,
                                     sharp corners=south,
                                     enhanced,
                                     breakable]
\scriptsize
\begin{verbatim}
Let $T=256$.
Let $\mathcal{R}$ be the region in the plane defined by the inequalities
$x^{2}+y^{2} \geq T$ and $|x|+|y| \leq \sqrt{2 T}$.
Compute the area of region $\mathcal{R}$.
\end{verbatim}
\end{tcolorbox}

\begin{tcolorbox}[title=Olympiad Bench Result for Qwen2.5-Math-7B-Instruct after SFT,
  colback=yellow!5,
  colframe=brown!80,
  fonttitle=\bfseries,
  sharp corners=south,
  enhanced,
  breakable]
  \scriptsize
\label{box:olympiad-sft-result}
\begin{verbatim}

**Problem:**  
Let \( T = 256 \). Define the region \(\mathcal{R}\) in the plane by the inequalities  
\[
x^{2} + y^{2} \geq T \quad \text{and} \quad |x| + |y| \leq \sqrt{2T}.
\]  
Compute the area of \(\mathcal{R}\).

---

### Step 1: Understand the region \(\mathcal{R}\)

- The inequality \( x^{2} + y^{2} \geq T \) describes the exterior of the circle centered at the
origin with radius \(\sqrt{T}\).
- The inequality \( |x| + |y| \leq \sqrt{2T} \) describes the interior of the square centered
at the origin with vertices at \((\pm \sqrt{2T}, 0)\) and \((0, \pm \sqrt{2T})\).

Since \( T = 256 \), we have  
\[
\sqrt{T} = \sqrt{256} = 16, \quad \sqrt{2T} = \sqrt{2 \times 256} = \sqrt{512} = 16 \sqrt{2}.
\]

Thus, the region \(\mathcal{R}\) is the area inside the square  
\[
|x| + |y| \leq 16 \sqrt{2}
\]  
but outside the circle  
\[
x^{2} + y^{2} \leq 16^{2} = 256.
\]

---

### Step 2: Calculate the area of the square

The square is defined by the inequality  
\[
|x| + |y| \leq 16 \sqrt{2}.
\]

The vertices of this square are at  
\[
(\pm 16 \sqrt{2}, 0) \quad \text{and} \quad (0, \pm 16 \sqrt{2}).
\]

The length of one side of the square is the distance between \((16 \sqrt{2}, 0)\) and \((0, 16 \sqrt{2})\).
    Using the distance formula:
\[
\text{side length} = \sqrt{(16 \sqrt{2} - 0)^2 + (0 - 16 \sqrt{2})^2}
    = \sqrt{(16 \sqrt{2})^2 + (16 \sqrt{2})^2} = \sqrt{2 \times (16 \sqrt{2})^2}.
\]

Calculate \((16 \sqrt{2})^2\):  
\[
(16 \sqrt{2})^2 = 16^2 \times 2 = 256 \times 2 = 512.
\]

Therefore,  
\[
\text{side length} = \sqrt{2 \times 512} = \sqrt{1024} = 32.
\]

The area of the square is  
\[
\text{side length}^2 = 32^2 = 1024.
\]

---

### Step 3: Calculate the area of the circle

The circle is defined by  
\[
x^{2} + y^{2} \leq 256,
\]  
with radius \( r = 16 \).

The area of the circle is  
\[
\pi r^{2} = \pi \times 16^{2} = 256 \pi.
\]

---

### Step 4: Calculate the area of the region \(\mathcal{R}\)

The region \(\mathcal{R}\) is the part of the square that lies outside the circle. Therefore, its area is  
\[
\text{Area}(\mathcal{R}) = \text{Area of square} - \text{Area of circle} = 1024 - 256 \pi.
\]

---

### **Final answer:**

\[
\boxed{1024 - 256 \pi}.
\]

\end{verbatim}
\end{tcolorbox}

\begin{tcolorbox}[title=Olympiad Bench Result for Qwen2.5-Math-7B-Instruct,
  colback=yellow!5,
  colframe=brown!80,
  fonttitle=\bfseries,
  sharp corners=south,
  enhanced,
  breakable]
  \scriptsize
\label{box:olympiad-qwen-result}
\begin{verbatim}
To find the area of the region \(\mathcal{R}\) defined by the inequalities \(x^2 + y^2 \geq T\) and 
\(|x| + |y| \leq \sqrt{2T}\), we start by substituting \(T = 256\). This transforms the 
inequalities into \(x^2 + y^2 \geq 256\) and \(|x| + |y| \leq 16\).

The inequality \(x^2 + y^2 \geq 256\) describes the region outside the circle centered at the 
origin with radius 16. The inequality \(|x| + |y| \leq 16\) describes a square centered at the 
origin with vertices at \((16,0)\), \((-16,0)\), \((0,16)\), and \((0,-16)\).

The region \(\mathcal{R}\) is the area inside the square but outside the circle. To find this area, 
we calculate the area of the square and subtract the area of the circle.

The area of the square is:
\[
16 \times 16 = 256
\]

The area of the circle is:
\[
\pi \times 16^2 = 256\pi
\]

The area of the region \(\mathcal{R}\) is:
\[
256 - 256\pi
\]

Thus, the area of region \(\mathcal{R}\) is:
\[
\boxed{256(1 - \pi)}
\]

\end{verbatim}
\end{tcolorbox}

\begin{tcolorbox}[title=GSM8K Sample Problem Definition,
  colback=yellow!5,
  colframe=brown!80,
  fonttitle=\bfseries,
  sharp corners=south,
  enhanced,
  breakable]
  \scriptsize
\label{box:gsm88k-problem}
\begin{verbatim}
A pink frog weighs the same as a blue beetle, the same as a green toad,
10 pounds less than a yellow snake, but 20 pounds more than a red bird.
If the pink frog weighs 50 pounds,
calculate the total weight of a container carrying one of each animal if the container also weighs 20 pounds.
\end{verbatim}
\end{tcolorbox}

\begin{tcolorbox}[title=GSM8K Result for Qwen2.5-Math-7B-Instruct after SFT,
  colback=yellow!5,
  colframe=brown!80,
  fonttitle=\bfseries,
  sharp corners=south,
  enhanced,
  breakable]
  \scriptsize
\begin{verbatim}
Let's analyze the problem step-by-step.

**Given:**
- The pink frog weighs 50 pounds.
- The pink frog weighs the same as the blue disgosition.
- The blue disgosition weighs the same as the green toad.
- The green toad weighs 10 pounds less than the yellow snake.
- The yellow snake weighs 20 pounds more than the red bird.
- The container weighs 20 pounds.

**Step 1: Define variables for the weights of each animal.**

Let:
- \( F \) = weight of the pink frog
- \( B \) = weight of the blue disgosition
- \( G \) = weight of the green toad
- \( S \) = weight of the yellow snake
- \( R \) = weight of the red bird

**Step 2: Write equations based on the given relationships.**

1. The pink frog weighs the same as the blue disgosition:
\[
F = B
\]

2. The blue disgosition weighs the same as the green toad:
\[
B = G
\]

3. The green toad weighs 10 pounds less than the yellow snake:
\[
G = S - 10
\]

4. The yellow snake weighs 20 pounds more than the red bird:
\[
S = R + 20
\]

5. The pink frog weighs 50 pounds:
\[
F = 50
\]

**Step 3: Use the known value to find the other weights.**

From \( F = 50 \), and since \( F = B \), we have:
\[
B = 50
\]

Since \( B = G \), we also have:
\[
G = 50
\]

Using \( G = S - 10 \), substitute \( G = 50 \):
\[
50 = S - 10 \implies S = 60
\]

Using \( S = R + 20 \), substitute \( S = 60 \):
\[
60 = R + 20 \implies R = 40
\]

**Step 4: Calculate the total weight of the container carrying one of each animal.**

Sum the weights of the pink frog, blue disgosition, green toad, yellow snake, red bird, and the container:
\[
F + B + G + S + R + \text{container} = 50 + 50 + 50 + 60 + 40 + 20
\]

Calculate the sum:
\[
50 + 50 = 100 \\
100 + 50 = 150 \\
150 + 60 = 210 \\
210 + 40 = 250 \\
250 + 20 = 270
\]

**Final answer:**

\[
\boxed{270}
\]

The total weight of the container carrying one pink frog, one blue disgosition,
one green toad, one yellow snake, one red bird, and the container itself is **270 pounds**.""
\end{verbatim}
\end{tcolorbox}

\begin{tcolorbox}[title=GSM8K Result for Qwen2.5-Math-7B-Instruct,
                  colback=yellow!5,
                  colframe=brown!80,
                  fonttitle=\bfseries,
                  sharp corners=south,
                  enhanced,
                  breakable]
\scriptsize
\label{box:gsm8k-qwen-result}
\begin{verbatim}
To determine the total weight of a container carrying one pink frog,
one blue beetle, one green toad, one yellow snake, and one red bird,
we need to find the weight of each animal based on the given relationships and
then sum these weights along with the weight of the container.

1. **Weight of the pink frog**: Given as 50 pounds.
2. **Weight of the blue beetle**: The blue beetle weighs the same as the pink frog, so it also weighs 50 pounds.
3. **Weight of the green toad**: The green toad weighs the same as the pink frog, so it also weighs 50 pounds.
4. **Weight of the yellow snake**: The yellow snake weighs 10 pounds more than the pink frog.
Therefore, the weight of the yellow snake is:
   \[
   50 + 10 = 60 \text{ pounds}
   \]
5. **Weight of the red bird**: The red bird weighs 20 pounds less than the pink frog.
Therefore, the weight of the red bird is:
   \[
   50 - 20 = 30 \text{ pounds}
   \]

Now, we sum the weights of all the animals and the container:
\[
50 \text{ (pink frog)} + 50 \text{ (blue beetle)} + 50 \text{ (green toad)} +
    60 \text{ (yellow snake)} + 30 \text{ (red bird)} + 20 \text{ (container)} = 260 \text{ pounds}
\]

Thus, the total weight of the container carrying one of each animal is \(\boxed{260}\).


\end{verbatim}
\end{tcolorbox}

\end{document}